%% file: arxiv.tex
\documentclass[conference]{IEEEtran}
\usepackage{times}

\usepackage[numbers]{natbib}
\usepackage{multicol}

\usepackage{graphicx} 
\usepackage{wrapfig}

\usepackage[export]{adjustbox}
\usepackage[position=top]{subfig}
\usepackage{subcaption} 

\usepackage{threeparttable}
\usepackage{makecell}
\usepackage{multirow}
\usepackage{booktabs} 

\usepackage{pifont}
\usepackage[ruled, lined, longend, linesnumbered]{algorithm2e}

\usepackage{amsmath}
\usepackage{amssymb}
\usepackage{amsfonts}

\usepackage{xcolor}

\definecolor{violet}{RGB}{148,0,211}

\usepackage[bookmarks=true]{hyperref}
\usepackage{url}

\usepackage{hhline}

\usepackage{pdfpages}

\usepackage{listings}   
\usepackage{placeins}   
\usepackage{tikz}
\usetikzlibrary{arrows.meta, positioning, shapes.geometric, calc}

\input{format}

\begin{document}


\title{\textbf{FLASH}: \textbf{F}ast \textbf{L}earning via GPU-\textbf{A}ccelerated \textbf{S}imulation for \textbf{H}igh-Fidelity Deformable Manipulation in Minutes}




%
\author{\authorblockN{Siyuan Luo$^{1}$\authorrefmark{2}\authorrefmark{1},
Bingyang Zhou$^{1}$\authorrefmark{1},
Chong Zhang$^{2}$\authorrefmark{1},
Xin Liu$^{1}$,
Zhenhao Huang$^{1}$,
Gang Yang$^{1}$,\\
Zhengtao Han$^{3}$,
Xiaotian Hu$^{4}$,
Eric Yang$^{5}$,
Rymon Yu$^{5}$,
Ziqiu Zeng$^{1}$,
and Fan Shi$^{1}$\authorrefmark{2}}
\authorblockA{$^{1}$NUS Human-Centered Robotic Lab \, $^{2}$ETH\, $^{3}$ShanghaiTech University\, $^{4}$Independent Researcher\, $^{5}$Gradient}
\authorblockA{\authorrefmark{1} Equal Contribution\,  \authorrefmark{2}Corresponding: \small{sy.luo@nus.edu.sg, fan.shi@nus.edu.sg}}\\
\authorblockA{Website: \href{https://siyuanluo.com/flash}{https://siyuanluo.com/flash}\, \\ \small{Submitted on Jan 30, 2026. First revision (current version) on Mar 23, 2026.} }
}

\maketitle

\begin{strip}
    \centering
    \includegraphics[width=0.95\linewidth]{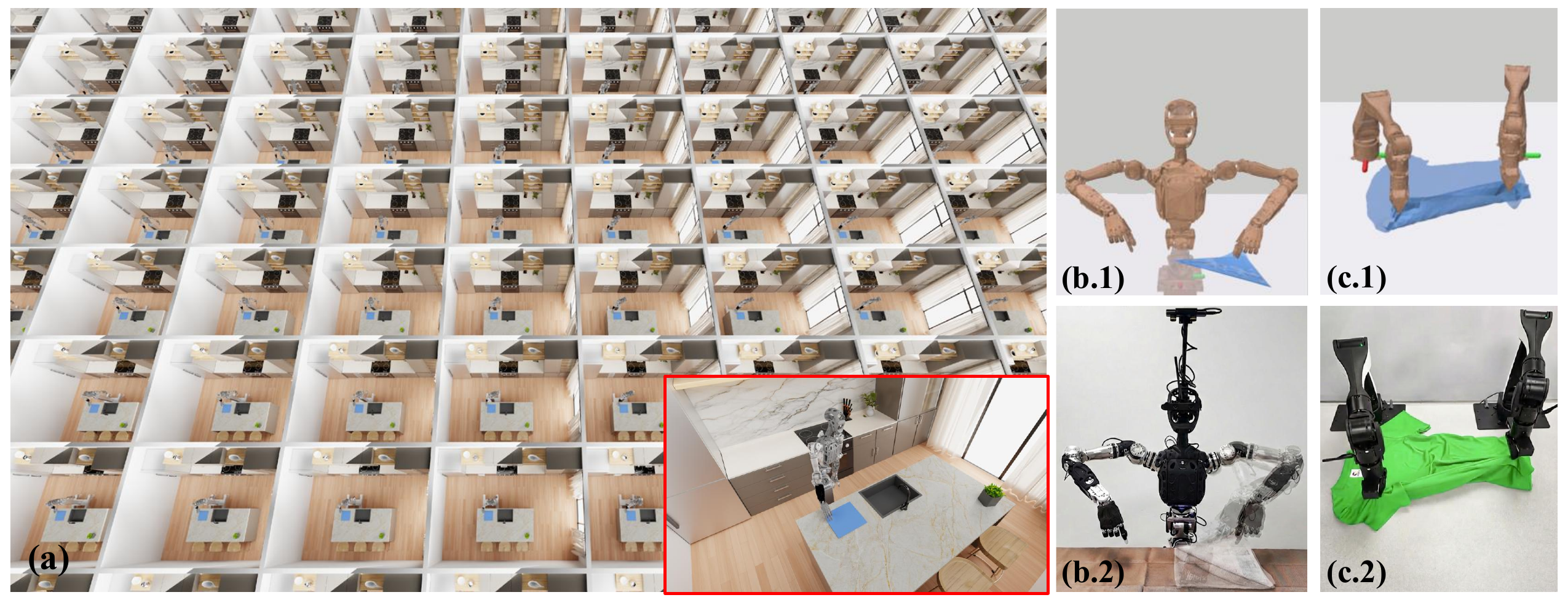} 
    \captionof{figure}{\textbf{Overview of the FLASH framework for GPU-parallel simulation and robot policy learning.} (a) A custom, GPU-native deformable simulator executes massively parallel rollouts; scalability is enabled by our physical solvers tailored for GPU-friendly parallelism. (b–c) Qualitative results of two deformable manipulation tasks in simulation and on real hardware, illustrating high-fidelity contact and deformation. FLASH achieves 100–300× end-to-end training speedup compared to realtime.}
    \label{Fig: teaser} 
\end{strip}

\input{sections/0_abstract}

\input{sections/1_introduction}

\input{sections/2_related_works}

\input{sections/3_prelim}
\input{sections/4_method}
\input{sections/5_experiments}
\input{sections/6_real_robot}
\input{sections/7_conclusion}

\IEEEpeerreviewmaketitle

\input{sections/10_acknowledgement}


\bibliographystyle{plainnat}
\bibliography{references}

\clearpage
\onecolumn
\input{sections/appendix}

\end{document}

%% file: format.tex
\usepackage{amsmath}
\usepackage{bm}

\usepackage[ruled]{algorithm2e} 

\SetKwComment{Comment}{$//$\ }{}

\usepackage{empheq} 
\SetKwComment{Comment}{$//$\ }{}
\usepackage{tikz,pgfplots} 
\pgfplotsset{width=\linewidth, compat=1.9}
\usepackage{subcaption}
\usepackage{mathtools}
\usepackage{multirow}
\usepackage{dblfloatfix}

\usepackage[export]{adjustbox}
\usepackage{afterpage}
\usepackage{placeins}
\usepackage{makecell}
\usepackage{cuted}
\usepackage{capt-of}
\usepackage{booktabs}
\usepackage{tabularx}  
\usepackage[table,xcdraw]{xcolor} 
\usepackage{pifont}  
\usepackage{array}
\usepackage{tabularx}
\usepackage{listings}
\usepackage[T1]{fontenc}
\usepackage{inconsolata}

\newcolumntype{M}{>{\centering\arraybackslash}m}
\newcolumntype{C}{>{\centering\arraybackslash}X} 

\newcommand{\yes}{\ding{51}}
\newcommand{\no}{\ding{55}}  

\newcommand{\zeros}{\mathbf{0}} 

\newcommand{\pos}{\mathbf{q}} 
\newcommand{\vel}{\mathbf{v}} 
\newcommand{\bpos}{\mathbf{q}_h} 
\newcommand{\bvel}{\mathbf{v}_h} 
\newcommand{\ppos}{\tilde{\mathbf{q}}} 
\newcommand{\ext}{\mathbf{f}_{\mathrm{ext}}} 
\newcommand{\internal}{\mathbf{f}_{\mathrm{int}}} 

\newcommand{\force}{\boldsymbol{\lambda}}

\newcommand{\proj}{\mathbf{p}}
\newcommand{\weight}{w_i}

\newcommand{\mass}{\mathbf{M}} 
\newcommand{\map}{\mathbf{G}_i} 
\newcommand{\sys}{\mathbf{A}} 
\newcommand{\jac}{\mathbf{H}} 
\newcommand{\nsjacobian}{\mathbf{J}} 

\newcommand{\lset}{\mathcal{L}} 

\newcommand{\inv}[1]{#1^{-1}}
\newcommand{\trans}[1]{#1^{\top}}

\usepackage{pifont}

\DeclareMathOperator{\blkdiag}{blkdiag}
\DeclareMathOperator{\col}{col}

\definecolor{codeblue}{rgb}{0.1, 0.35, 0.65}   
\definecolor{codegreen}{rgb}{0.15, 0.5, 0.25}  
\definecolor{codepurple}{rgb}{0.6, 0.1, 0.5}   

\lstdefinestyle{academic}{
    backgroundcolor=\color{white},   
    commentstyle=\color{codegreen}\itshape,
    keywordstyle=\color{codeblue}\bfseries,
    stringstyle=\color{codepurple},
    basicstyle=\ttfamily\footnotesize,
    breakatwhitespace=false,         
    breaklines=true,                 
    captionpos=b,                    
    keepspaces=true,                 
    numbers=none,
    showspaces=false,                
    showstringspaces=false,
    showtabs=false,                  
    tabsize=4,
    frame=tb,                        
    framerule=0.5pt,                 
    xleftmargin=0pt,
}

%% file: sections/0_abstract.tex
\begin{abstract}
Simulation frameworks such as Isaac Sim have enabled scalable robot learning for locomotion and rigid-body manipulation; however, contact-rich simulation remains a major bottleneck for deformable object manipulation. The continuously changing geometry of soft materials, together with large numbers of vertices and contact constraints, makes it difficult to achieve high accuracy, speed, and stability required for large-scale interactive learning. We present FLASH, a GPU-native simulation framework for contact-rich deformable manipulation, built on an accurate NCP-based solver that enforces strict contact and deformation constraints while being explicitly designed for fine-grained GPU parallelism. Rather than porting conventional single-instruction-multiple-data (SIMD) solvers to GPUs, FLASH redesigns the physics engine from the ground up to leverage modern GPU architectures, including optimized collision handling and memory layouts. As a result, FLASH scales to over 3 million degrees of freedom at 30 FPS on a single RTX 5090, while accurately simulating physical interactions. Policies trained solely on FLASH-generated synthetic data in minutes achieve robust zero-shot sim-to-real transfer, which we validate on physical robots performing challenging deformable manipulation tasks such as towel folding and garment folding, without any real-world demonstration, providing a practical alternative to labor-intensive real-world data collection.

\end{abstract}

%% file: sections/1_introduction.tex
\section{Introduction}~\label{Sec: Intro}
Deformable object manipulation is a long-standing challenge in robotics, with broad applications in domestic services, healthcare, and industrial automation. Unlike rigid bodies, deformable objects exhibit high degrees of freedom and complex dynamics, making generalized manipulation exceptionally difficult. Learning-based methods offer a promising direction, but they are inherently data-hungry, often requires millions of environment interactions to cover space of possible behaviors. Collecting such large-scale interaction data in the real world is inefficient and costly, and may provide limited physical diversity for robust policy generalization. This places significant demands on the underlying simulation platform: it must simultaneously delivery high physical fidelity and high computational throughput. 

\newcolumntype{Y}{>{\centering\arraybackslash}X}

\begin{table*}[t]
\centering
\caption{Comparison of Robotics Simulation Platforms}
\label{table:sim_comparison}
\renewcommand{\arraystretch}{1.2}
\begin{tabularx}{\textwidth}{c|c|Y|c|c|c|Y|c}
\toprule
\rowcolor[HTML]{EFEFEF} 
\textbf{Platform} & \textbf{Deformable} & \textbf{Dynamic Solver} & \textbf{Contact Model} & \textbf{Multi-Envs Training} & \textbf{GPU} & \textbf{Torch Support} & \textbf{Realistic Render} \\ \midrule
Issac Sim\cite{makoviychuk2021isaac} & \yes & XPBD, FEM & Compliant & \yes & \yes & \yes & \yes \\ \hline
Newton\cite{Newton_Physics_2025} & \yes & XPBD, VBD & Compliant & \yes & \yes & \no & \no \\ \hline
Physx\cite{macklin2019physx} & \yes & XPBD, FEM & Compliant & \no & \yes & \no & \no \\ \hline
Genesis\cite{Genesis} & \yes & PBD, MPM & Compliant & \yes & \yes & \yes & \no \\ \hline
SOFA\cite{westwood2007sofa} & \yes & FEM & LCP-PGS\tnote{a} & \no & \yes & \no & \no \\ \hline
Pinocchio\cite{carpentier2019pinocchio} & \no & \textcolor{gray}{\textbf{/}} & NCP (rigid) & \no & \no & \no & \no \\ \hline
Mujoco\cite{todorov2012mujoco} & \no & \textcolor{gray}{\textbf{/}} & CCP\tnote{a} (rigid) & \yes & \yes & \yes & \no \\ \hline
\rowcolor[HTML]{C7F0EE} 
\textbf{Flash (Ours)} & \textbf{\yes} & \textbf{FEM} & \textbf{NCP} & \textbf{\yes} & \textbf{\yes} & \yes & \textbf{\yes} \\ \bottomrule
\end{tabularx}
\begin{tablenotes}
\footnotesize
\item[a] LCP/CCP/NCP: Linear/Cone/Nonlinear Complementarity Problem; PGS: Projected Gauss-Seidel method.
\end{tablenotes}
\vspace{-10pt}
\end{table*}

Existing simulators, however, struggle to satisfy both requirements. CPU-based high-precision physics engines lack scalability for large-scale parallel learning, while prior GPU-based approaches are often limited by architectural constraints such as insufficient solver-level parallelism and inefficient memory access patterns, failing to fully exploit GPU computational potential. Consequently, developing a high-performance numerical framework tailored for deformable object simulation that is explicitly designed to leverage GPU parallelism is a critical prerequisite for scalable robot learning in deformable manipulation.

To address these challenges, we introduce \textbf{FLASH}, a GPU-friendly framework for high-performance simulation, rendering, and learning, enabling simulation-only training of deployable deformable manipulation policies. The key insight behind FLASH is that scalable learning for deformable manipulation requires solver-level and rendering-level designs that are co-optimized for GPU parallelism, rather than treating simulation and learning as separate components. At the simulation level, FLASH models deformable dynamics with coupled constraints and introduces a lightweight non-smooth Newton contact solver designed for massive GPU parallelism; a local–global strategy further improves robustness under long-horizon time steps. At the rendering level, FLASH tightly couples depth rendering and occlusion handling with simulation to minimize data movement and latency. At the learning level, FLASH exploits large-scale, diverse simulated interactions, generated through systematic physical randomization and state-based supervision, to train vision-based manipulation policies without real-world demonstrations. Together, these design principles bridge high-fidelity deformable physics and scalable policy learning, enabling robust sim-to-real transfer on contact-rich manipulation tasks.

By unifying high-fidelity simulation, efficient rendering, and scalable policy learning, we make the following contributions:
\begin{enumerate}
    \item \textbf{High-Performance Deformable Simulation and Rendering.} A GPU-optimized architecture built from the ground up, featuring with a custom non-smooth Newton solver achieves real-time (30fps) simulation of 3M+ DoF contact-rich deformable scenes, achieving 100-300$\times$ reduction in clock time on a single consumer GPU.
    
    \item \textbf{High-Throughput Synthesized Data Training.} A fully automated pipeline learns vision-based manipulation policies at scale without real-world demonstrations, leveraging massive simulated interactions, systematic domain randomization, and state-based supervision.
    
    \item \textbf{Zero-Shot Sim-to-Real Transfer.} High-fidelity contact modeling and occlusion-aware depth rendering enable policies trained entirely in simulation to transfer directly to real robots, operating robustly under perceptual uncertainty.
    
    \item \textbf{Generality Across Objects and Settings.} General framework supports multiple deformable objects (cloth and volumetric materials) and learning settings without task-specific simulator redesign.
    
\end{enumerate}

%% file: sections/2_related_works.tex
\section{Related Works}~\label{Sec: Related Works}
\vspace{-0.5\baselineskip}
\subsection{Deformable Objects Simulation}
Physics simulation for deformable objects has evolved along two main directions. The first focuses on accuracy, exemplified by SOFA~\cite{westwood2007sofa}, which provides high-fidelity soft tissue modeling for medical applications. The second prioritizes speed for robot learning. Early efforts like MuJoCo~\cite{todorov2012mujoco} offered efficient rigid-body simulation but limited soft-body support. Recent GPU-accelerated platforms have made large-scale deformable simulation practical: Isaac Sim~\cite{makoviychuk2021isaac} uses Position-based Dynamics (PBD) on GPUs for parallel policy training, Genesis~\cite{Genesis} combines Material-Point Method (MPM)~\cite{stomakhin2013material} and PBD solvers for diverse soft materials, and Newton~\cite{Newton_Physics_2025} provides a unified differentiable framework with multi-physics coupling, including Projective Dynamics (PD) and Vertex Block Descent (VBD)~\cite{chen2024vertex}. Despite these advances, existing simulators still struggle to balance physical fidelity with the throughput needed for contact-rich manipulation learning.

The performance of these simulators largely depends on their underlying numerical methods. The FEM~\cite{sifakis2012fem} offers high accuracy by discretizing continuum mechanics equations, but its computational cost limits real-time applications. To address this, PBD~\cite{muller2007position} takes a different approach by directly updating positions through constraint projection, achieving faster and more stable simulation; however, its effective stiffness depends on iteration number, making it difficult to model stiff materials. Extended PBD (XPBD)~\cite{macklin2016xpbd} addresses this limitation by introducing a compliance parameter that decouples stiffness from solver iterations. PD~\cite{zeng2025fast} offers another solution, combining FEM's physical accuracy with PBD's efficiency through a local-global optimization framework.

\subsection{Sim-to-Real Learning of Deformable Manipulation} 
Sim-to-real learning offers a scalable alternative to costly real-world data collection~\cite{bu2025agibot, o2024open, zitkovich2023rt}.
Early works demonstrated that domain randomization~\cite{pmlr-v87-matas18a, clegg2020learning} or algorithmic supervisors~\cite{seita2020deep} could enable policies trained in simulation to transfer to the real world without real-world demonstrations. To facilitate efficient learning from simulated physical interactions, many approaches simplify the control problem by restricting actions to robust primitives (such as quasi-static picking~\cite{wu2020learning}, flinging~\cite{ha2022flingbot}, or velocity-parameterized motions~\cite{blanco2023qdp}) or use primitives to generate data for imitation learning~\cite{wang2025dexgarmentlab, lu2024garmentlab}. To tackle the complexity of manipulation tasks, prior work has also explored alternative formulations, such as predicting dense deformation flows~\cite{weng2022fabricflownet}, utilizing generative dynamics models~\cite{tian2025diffusion}, or integrating language-guided trajectory planners~\cite{chen2025metafold, barbany2025bifold}. To guide exploration in high-dimensional state spaces, some methods also rely on expert demonstrations~\cite{salhotra2022learning}.

However, a significant sim-to-real gap remains in both dynamics and perception. Recent benchmarks~\cite{ru2025can, hu2025real} indicate that standard simulators struggle to faithfully reproduce dynamic fabric behaviors, often requiring complex real-to-sim parameter tuning~\cite{manyar2025real} or learning residual delta dynamics to iteratively refine trajectories~\cite{chi2024iterative}. Regarding perception, researchers bridge the perception gap by either abstracting away appearance discrepancies using semantic keypoints~\cite{lips2024learning, deng2025general} or enhancing simulation fidelity via mesh-attached neural fields~\cite{delehelle2024garfield}.

These efforts collectively show that (i) learning directly from simulation can transfer, but (ii) performance is often bounded by the simulator constraints: not only fidelity, but also runtime, stability, and scalability. Our work addresses this with fast and realistic simulation, enabling high-throughput learning of deployable policies directly from simulation with contact-rich physical interactions, and supporting broad domain randomization and generation of diverse recovery scenarios to improve real-world robustness.

%% file: sections/3_prelim.tex
\section{Preliminary}~\label{Sec: Prelim}

Our simulation-learning framework builds upon the GPU-friendly local-global pipeline proposed in~\cite{zeng2025fast}. 
Given positions $\bpos$ and velocities $\bvel$ at time $t$, the implicit Euler integration computes the updated positions $\pos$ at time $t+h$ by solving
\begin{equation}
    \pos = \arg\min_{\pos'} \left(
    \frac{1}{2h^2} \big\|\mass^{\frac{1}{2}}(\pos'-\ppos)\big\|^2_F 
    + \sum\nolimits_i \psi_i(\pos')
    \right),
    \label{eq:implicit}
\end{equation}
where $\mass$ is the mass matrix and $\ppos = \bpos + h\bvel + h^2 \inv{\mass}\ext$ denotes the predicted state. 
The function $\psi_i$ represents the elastic energy associated with the $i$-th finite element.

\begin{figure*}[t]
    \centering
    \includegraphics[trim={0.5cm 0 0.5cm 0.8cm},clip,width=0.9\linewidth]{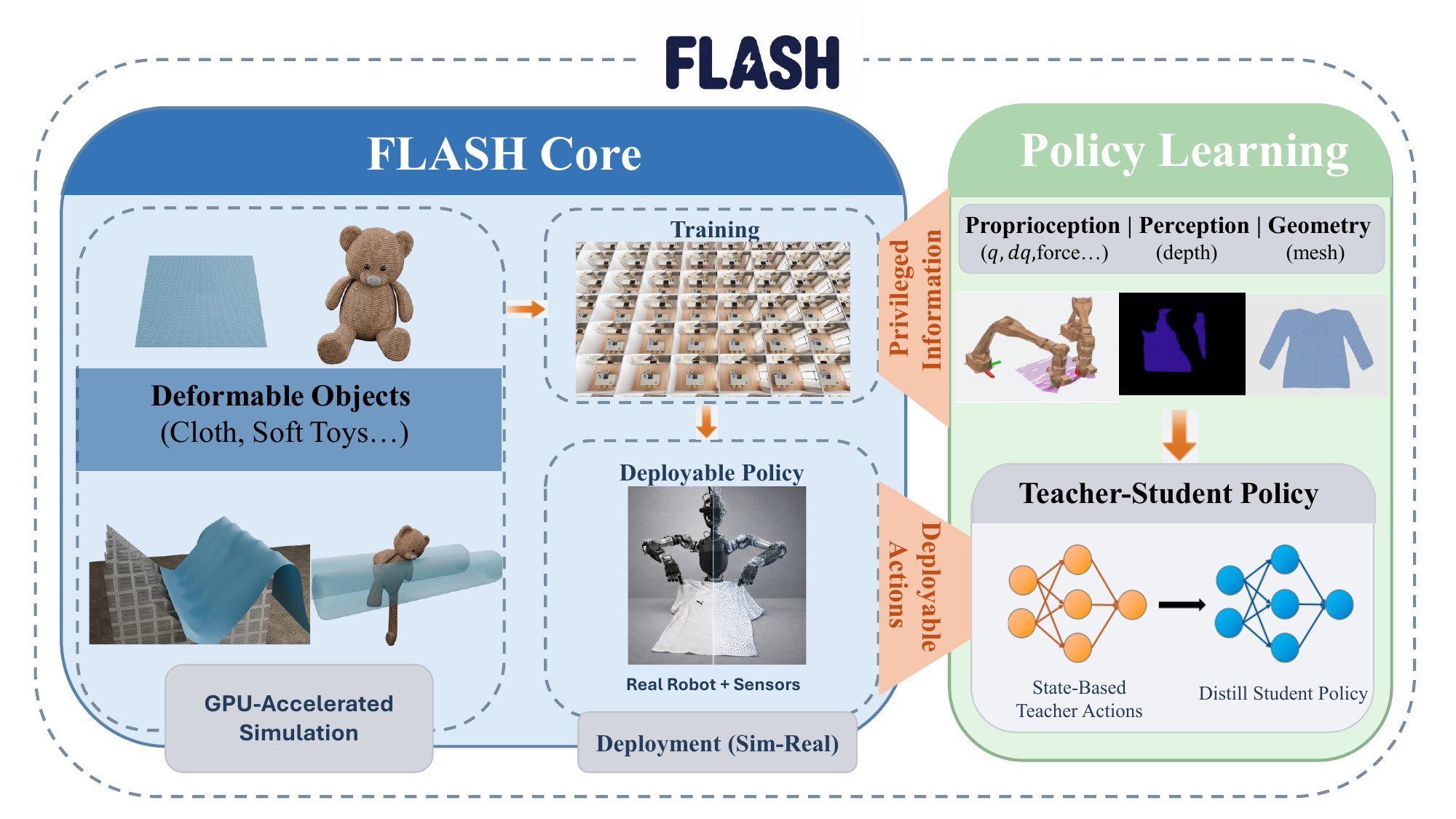}
    \caption{\textbf{An illustration of the FLASH System.} FLASH is
a GPU-accelerated high-fidelity deformable simulation framework that enables fast, large-scale learning for robotic manipulation. On top of FLASH, we build a learning and control pipeline that constructs observations from proprioception and perception, trains deployable policies via imitation learning, and transfers the learned policy to real robots for deformable manipulation tasks. }
    \label{Fig: pipeline}
    \vspace{-20pt}
\end{figure*}

\subsection{Contact-Constrained Dynamics}
To account for contact interactions, the Lagrange multiplier formulation~\cite{macklin2019non} augments the dynamics by introducing contact forces $\force$ as constraint multipliers. 
Differentiating the objective in Eq.\eqref{eq:implicit} under contact constraints yields the following system:
\begin{subequations}
\begin{align}
    \mass(\pos-\ppos) - h^2 \internal(\pos) - h^2 \sum_{j\in\lset} \trans{\jac}_j \force &= \zeros \label{eq:lagrange_a}\\
    \forall j\in\lset, \quad \boldsymbol{\phi}_j(\pos, \force) &= \zeros, \label{eq:lagrange_b}
\end{align}
\label{eq:lagrange}
\end{subequations}
where $\internal$ denotes implicit internal forces derived from the elastic energies, and $\lset$ is the set of all Lagrangian constraints. 
The contact Jacobian $\jac$ maps the generalized coordinates to the contact space, while $\boldsymbol{\phi}_j$ encodes either bilateral conditions or Signorini--Coulomb contact laws, depending on the constraint type.

\subsection{Projective Internal Forces}
Following the projective dynamics formulation \cite{bouaziz2023projective, overby_admm_2017}, internal elastic forces are evaluated via auxiliary projective states.
For each local elastic constraint, a projective variable $\proj_i$ is obtained by solving a proximal problem:
\begin{subequations}
\begin{align}
    \proj_i = \arg\min_{\proj'_i} 
    \left(
    \frac{\weight}{2} \|\proj'_i - \map \pos\|_F^2 + \zeta_i(\proj'_i, \xi_i)
    \right)& \\
    \internal(\pos) = \sum\nolimits_i \weight \trans{\map} \big(\proj_i - \map \pos\big)&,
\end{align}
\label{eq:internal}
\end{subequations}
where $\weight \ge 0$ is the stiffness weight of the $i$-th constraint, $\map$ is the linear mapping from the mechanical state to the projection state, and $\zeta_i$ defines the constitutive manifold through its minimizers.

Substituting Eq.\eqref{eq:internal} into Eq.\eqref{eq:lagrange} and assembling all constraints, the deformable multi-body system can be written as
\begin{subequations}
\begin{align}
    \sys \pos - \mathbf{b} - h^2 \trans{\jac} \force &= \zeros \label{eq:dynamics_a}\\
    \boldsymbol{\phi}(\pos, \force) &= \zeros, \label{eq:dynamics_b}
\end{align}
\label{eq:dynamics}
\end{subequations}
where the system matrix and right-hand side are defined as
\[
\sys = \mass + h^2 \sum\nolimits_i \weight \trans{\map}\map,
\qquad
\mathbf{b} = \mass \ppos + h^2 \sum\nolimits_i \weight \trans{\map}\proj_i.
\]

\subsection{Non-Smooth Newton System}
To solve Eq.\eqref{eq:dynamics}, a non-smooth Newton method is employed.
Let
\begin{subequations}
\begin{align}
    \nsjacobian &= \frac{\partial \boldsymbol{\phi}}{\partial \pos}, \qquad
     \mathbf{g} = \mathbf{b} + h^2 \trans{\nsjacobian} \force^{k-1},\\
     \mathbf{E} &= \frac{\partial \boldsymbol{\phi}}{\partial \force},\qquad
    \mathbf{h} = -\boldsymbol{\phi}(\pos^{k-1}, \force^{k-1}) + \nsjacobian \pos^{k-1},
\end{align}
\end{subequations}
then each Newton iteration solves the saddle-point system
\begin{equation}
    \begin{bmatrix}
        \sys & -\nsjacobian^\top \\
        \nsjacobian & \mathbf{E}
    \end{bmatrix}
    \begin{bmatrix}
        \pos \\ h^2 \Delta \force
    \end{bmatrix}
    =
    \begin{bmatrix}
        \mathbf{g} \\ \mathbf{h}
    \end{bmatrix}.
    \label{eq:matrix_form}
\end{equation}

Solving Eq.\eqref{eq:matrix_form} within each local-global iteration involves three stages:
(i) assembling $\nsjacobian$, $\mathbf{E}$, $\mathbf{g}$, and $\mathbf{h}$, and forming the Schur complement 
\begin{equation}
    \mathbf{Z} = \nsjacobian \inv{\sys} \trans{\nsjacobian} + \mathbf{E};
    \label{eq:schur}
\end{equation}
(ii) solving for the constraint increment
\begin{equation}
    \Delta \force = \frac{1}{h^2}
    \inv{\mathbf{Z}}
    \big( \nsjacobian \inv{\sys} \mathbf{g} - \mathbf{h} \big);
    \label{eq:constraint_solve}
\end{equation}
and (iii) updating positions with constraint corrections
\begin{equation}
    \pos = \inv{\sys} \big( \mathbf{b} + h^2 \trans{\nsjacobian} (\force^{k-1} + \Delta \force) \big).
    \label{eq:correction}
\end{equation}

A key observation in~\cite{zeng2025fast} is that the system inverse admits a sparse factorization
$\inv{\sys} = \trans{\mathbf{S}} \mathbf{S}$,
where $\mathbf{S} = \inv{\mathbf{L}}$ is the explicit inverse of the Cholesky factor of $\sys$.
This representation enables both global linear solves and Schur-complement evaluations to be expressed as sparse matrix multiplications, making the overall pipeline highly amenable to GPU acceleration.

%% file: sections/4_method.tex
\section{FLASH SYSTEM}~\label{Sec: Method}
\vspace{-\baselineskip}
\subsection{System Overview}
In this section, we present the construction of FLASH, our simulation and learning platform specifically designed for contact-rich deformable manipulation. We separate this section into three parts. First, we describe the physics simulation core, including our non-smooth Newton solver, multi-environment parallelization and its GPU optimization (Sec.~\ref{Subsec: simulation}). Second, we explain the details of large-scale fast training (Sec.~\ref{Subsec: learning}). Finally, we present the real-to-sim transfer(Sec.~\ref{Subsec: sim2real}).

\subsection{Simulation Core}~\label{Subsec: simulation}
\vspace{-0.1\baselineskip}
The formulation described in Sec.~\ref{Sec: Prelim} provides the computational foundation of a high-performance hyperelastic simulation framework with friction-accurate contact, which has been proven effective for real-time deformable dynamics~\cite{zeng2025fast}.
In this work, we aim to build upon this framework to develop an integrated soft-body and robotic manipulation platform that supports contact-rich interaction, parallel simulation and learning, and scalable real-to-sim transfer.

While the original method~\cite{zeng2025fast}, demonstrates excellent performance for general deformable simulation, applying it to robotic manipulation introduces additional challenges. Many soft-material manipulation tasks (e.g., cloth folding and sliding contact) produce rich contact with time-dependent behaviors. 
Such contact-rich configurations activate a large number of constraints, significantly expanding the dimensionality of the Lagrange multiplier space and increasing both the computational cost and numerical difficulty of constraint resolution.

As the system formulated, the inverse of the system matrix $\inv{\sys}$ is, in general, fully dense. 
Consequently, the associated Schur-complement $\mathbf{Z} = \nsjacobian \inv{\sys} \trans{\nsjacobian} + \mathbf{E}$ also becomes fully dense in contact-dominated regimes. 
This density leads to a considerable computational burden when building Eq.\eqref{eq:schur} and solving Eq.\eqref{eq:constraint_solve} the constrained system, which becomes a major performance bottleneck in large-scale contact scenarios.
The issue is further exacerbated in parallelized simulation settings commonly used for robot learning, where multiple environments are simulated concurrently. 
In such cases, the cost of contact resolution scales unfavorably with both the number of environments and the number of active contact constraints, severely limiting throughput and practical scalability.

\subsubsection{A Lightweight Simulation System}
Motivated by these observations, we propose a lightweight variant of the method
in~\cite{zeng2025fast} that improves the sparsity of the Schur complement while
preserving numerical robustness.
Specifically, we approximate
\begin{equation}
\mathbf{Z} = \nsjacobian \inv{\sys} \trans{\nsjacobian} + \mathbf{E}
\;\approx\;
\nsjacobian \inv{\mass} \trans{\nsjacobian} + \mathbf{E}.
\end{equation}

This reformulation relaxes the fully implicit contact metric by adopting an
inertia-dominated inverse, reducing global contact coupling while retaining the
implicit treatment of elastic forces through projective dynamics.
Similar mass-based or inertia-weighted contact metrics have been widely adopted
in rigid-body and deformable contact formulations
\cite{stewart_implicit_1996, kaufman_staggered_2008} to improve sparsity and
scalability.
In contact-rich manipulation scenarios involving soft materials such as cloth or
foam-like objects, elastic coupling typically dominates system behavior, whereas
contact responses are largely governed by inertia and local geometry.
Under these conditions, the proposed approximation provides a favorable balance
between physical fidelity and computational efficiency.


Crucially, the resulting sparsity is preserved throughout the non-smooth Newton
iterations, enabling efficient and scalable simulation across multiple
environments, as discussed next.
For tasks that involve a small number of contacts but require strong global
contact coupling, the original fully implicit formulation can be retained,
allowing the contact metric to be selected flexibly based on task requirements.

\subsubsection{Multi-Env Supporting}

\begin{algorithm}[tb!]
\caption{Multi-Env GPU-based Pipeline}
\label{algo:lite}
\SetAlgoLined

Assemble block-diagonal operators $\bar{\sys}$\;
$\bar{\mathbf{L}} = \textit{Cholesky}(\bar{\sys}) ,\quad \bar{\mathbf{S}} = \inv{\bar{\mathbf{L}}}$\;

\While{\textit{simulation}}{

Collision detection (parallel across envs)\;

$\bar{\ppos} = \bar{\pos}_t + h\,\bar{\vel}_t + h^2 \inv{\bar{\mass}}\,\bar{\ext}$\;

\For{$k \in \{1,\dots,k_{\max}\}$}{

$\bar{\proj} = project(\bar{\map}\,\bar{\pos}^{k-1})$\Comment*[r]{local projection, batched over envs}

Build $\bar{\nsjacobian}$, $\bar{\mathbf{E}}$, $\bar{\mathbf{g}}$, $\bar{\mathbf{h}}$ from $\bar{\pos}^{k-1}$, $\bar{\force}^{k-1}$, and $\bar{\proj}$\;

$\bar{\mathbf{Z}} = \bar{\nsjacobian}\,\inv{\bar{\mass}}\,\bar{\nsjacobian}^\top + \bar{\mathbf{E}}$\;

$\Delta \bar{\force}
= \frac{1}{h^2}\,\inv{\bar{\mathbf{Z}}}\,
\Big(\bar{\mathbf{h}} - \bar{\nsjacobian}\,\bar{\mathbf{S}}^\top \bar{\mathbf{S}}\,\bar{\mathbf{g}}\Big)$\;

$\bar{\force}^{k} = \bar{\force}^{k-1} + \Delta \bar{\force}$\;

$\bar{\pos}^{k}
= \bar{\mathbf{S}}^\top \bar{\mathbf{S}}
\Big(\bar{\mathbf{b}} + h^2 \bar{\nsjacobian}^\top \bar{\force}^{k}\Big)$\;
}

$\bar{\pos}_{t+h} = \bar{\pos}^{k_{\max}},\quad
\bar{\vel}_{t+h} = \frac{1}{h}\big(\bar{\pos}_{t+h} - \bar{\pos}_t\big)$\;
}
\end{algorithm}

We first make the parallel structure across environments explicit by defining the block-diagonal system matrix and the stacked primal variable
\begin{equation}
\bar{\sys}
=
\begin{bmatrix}
\sys_{1} &        &        \\
         & \ddots &        \\
         &        & \sys_{n}
\end{bmatrix},
\qquad
\bar{\pos}
=
\begin{bmatrix}
\pos_{1} \\
\vdots \\
\pos_{n}
\end{bmatrix},
\label{eq:blockA}
\end{equation}
where $\sys_i$ and $\pos_i$ denote the system matrix and the primal variable associated with the $i$-th environment, respectively.
Analogously, all Jacobian- and system-related matrices are assembled in a block-diagonal manner across environments, while vector-valued quantities are stacked by concatenation.
For example, $\bar{\nsjacobian}=\blkdiag(\nsjacobian_1,\dots,\nsjacobian_n)$ and $\bar{\force}=\col(\force_1,\dots,\force_n)$.

Using the block-diagonal and stacked notation defined above, the non-smooth Newton system in Sec.~\ref{Sec: Prelim} can be directly extended to parallel environments.
Specifically, the saddle-point system in Eq.\eqref{eq:matrix_form} is assembled as
\begin{equation}
    \begin{bmatrix}
        \bar{\sys} & -\bar{\nsjacobian}^\top \\
        \bar{\nsjacobian} & \bar{\mathbf{E}}
    \end{bmatrix}
    \begin{bmatrix}
        \bar{\pos} \\ h^2 \Delta \bar{\force}
    \end{bmatrix}
    =
    \begin{bmatrix}
        \bar{\mathbf{g}} \\ \bar{\mathbf{h}}
    \end{bmatrix}.
    \label{eq:matrix_form_multienv}
\end{equation}
The corresponding Schur-complement system, constraint linear system, and position corrections follow identically from the single-environment formulation (Equations~\eqref{eq:schur}--\eqref{eq:correction}) by replacing scalar quantities with their block-diagonal or stacked counterparts.
Algorithm\ref{algo:lite} summarizes the resulting lightweight simulation pipeline and its multi-environment implementation.

\subsubsection{System-Level Parallelism}
Owning to the proposed inertia-dominated approximation, all stages of the resulting simulation pipeline operate on sparse and block-structured linear systems.
In particular, the dense coupling induced by $\inv{\sys}$ in the fully implicit contact formulation is eliminated, ensuring that the system matrix, Schur-complement, and associated linear operators remain sparse throughout the non-smooth Newton iterations. This property enables multiple simulation environments to be seamlessly assembled into a single large-scale sparse system using the block-diagonal and stacked representations defined above.
Importantly, no additional coupling is introduced beyond the block-diagonal assembly, allowing each environment to be processed independently within the same unified system.
All core operations are expressed using standard sparse primitives, allowing hardware parallelism, load balancing, and scheduling to be handled automatically by the underlying numerical libraries.

As a result, scalability is achieved at the system level rather than through manual parallelization.
The computational cost scales linearly with the number of environments and active contact constraints, providing a robust foundation for large-scale contact-rich simulation and parallel robot learning.

\subsubsection{Real-to-Sim Transfer and Rendering}
To bridge the gap between simulation and reality, we construct a high-fidelity digital twin focusing on geometry, physics, and visual alignment. 
\textbf{Geometric Modeling:} Mesh models for deformable objects are generated based on complexity: simple objects (e.g., square towels) are created procedurally, while complex garments (e.g., T-shirts) are reconstructed from 3D scans or from existing synthetic datasets~\cite{Zhou_2023_ICCV}. 
\textbf{Physical Calibration:} We tune key simulation parameters, such as Poisson's ratio and Young's modulus, to align the simulated deformation dynamics with real-world physical behaviors. 
\textbf{Visual Rendering:} Leveraging standard hand-eye calibration, we generate simulated depth images with robot self-occlusion that are spatially aligned with real-world data. To further account for unmodeled discrepancies, we apply domain randomization to both physical dynamics and visual observations during policy learning. Details in appendix.

\begin{figure}[th]
    \centering
    \includegraphics[width=0.8\columnwidth]{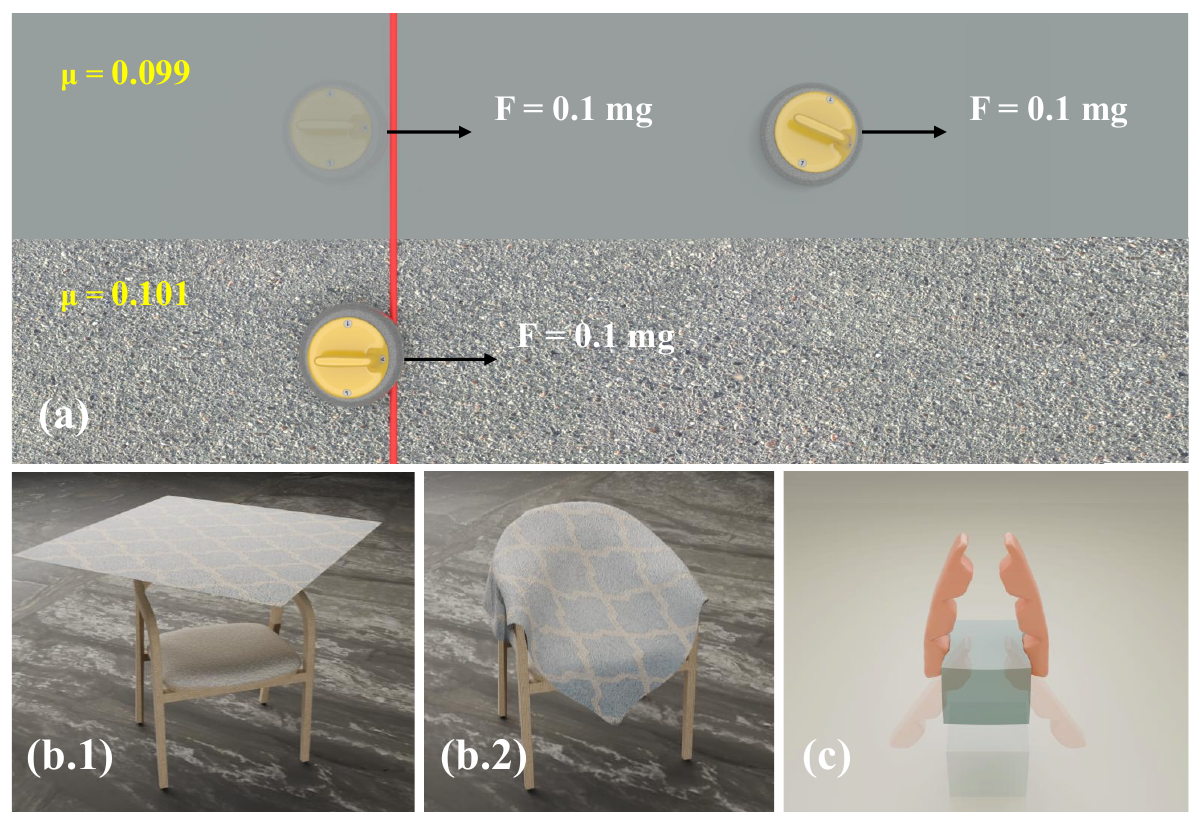}
    \caption{\textbf{Single environment simulation results.} We performed simulation experiments within a single environment, (a) Friction validation demonstrating the simulator's ability to distinguish coefficients with 0.001 precision. (b) Cloth simulation fidelity tested via cloth-chair collision. (c) A soft gripper grasping a soft cube via friction.}
    \label{Fig: final_sim}
    \vspace{-\baselineskip}
\end{figure}

\subsection{Policy Learning}~\label{Subsec: learning}
We adopt a teacher-student distillation framework~\cite{miki2022learning}, as illustrated in Fig.~\ref{Fig: pipeline}. 
We first synthesize teachers using privileged state information and heuristic rules that command end-effectors to grasp and transport specific keypoints to target locations. These heuristics also include reactive recovery behaviors to handle grasp failures. For complex tasks like garment folding, we organize these primitives into multi-stage finite-state machines. 
Subsequently, we distill these behaviors into deployable student policies, which are conditioned on a stacked history of proprioceptive and visual observations. To ensure sim-to-real robustness, we apply extensive domain randomization during this distillation process. Further technical details are provided in the supplementary materials.


\subsection{Sim-to-Real Transfer}~\label{Subsec: sim2real}
We achieve zero-shot sim-to-real transfer by explicitly aligning the observation and action spaces between simulation and the physical world.
The policy trained in simulation is directly deployed on the real robot without any fine-tuning.

\textbf{Perception.} 
We employ YOLOv8~\cite{jocher2023yolo} and SAM~\cite{ravi2024sam} to segment the target object from raw depth streams. 
This pipeline isolates the object of interest and filters out background noise, ensuring the real-world visual input matches the clean observations used in simulation.

\textbf{Control.} 
The policy predicts end-effector actions $a_t = (\Delta p, w)$ per arm, where $\Delta p \in \mathbb{R}^3$ denotes the position delta. 
The continuous gripper parameter $w \in \mathbb{R}$ is discretized into a binary open/close command via a sigmoid function. 
Integrating the position delta yields the Cartesian targets, which are converted into joint positions using a numerical Inverse Kinematics (IK) solver and executed by a high-frequency low-level controller.
For the dual-arm setup, the policy outputs 8-dimensional actions partitioned between the two arms.
Further details are provided in the supplementary material.

%% file: sections/5_experiments.tex
\section{Simulator \& Learning Results}~\label{Sec: Results}
We simulate cloth and volumes with frictional contact in Fig.~\ref{Fig: final_sim}. Then we evaluate high-fidelity cloth simulation on a challenging yet controlled \textbf{T-shirt dual-sleeve folding} task that closely mirrors real-world manipulation.
A single-layer T-shirt rests on a planar table with Coulomb friction ($\mu=1.0$), and cloth self-collision is disabled to ensure fair cross-platform comparison.
Two rigid cubic grippers are attached to the sleeve tips and driven in open-loop along identical Cartesian trajectories across all simulators: the left sleeve is folded across the midline and released, followed by a symmetric motion of the right sleeve.
All experiments are executed on a single RTX 4090 GPU.
Unless noted, gripper motions are slow and quasi-static (each fold lasts $5\mathrm{s}$ with $\Delta t = 0.01,\mathrm{s}$), ensuring that differences in the final garment state arise solely from each simulator’s handling of contact, friction, material response, and numerical stability rather than control or dynamic effects.

\subsection{Cross-simulator comparison}
\label{Subsec:cross_simulator}

\begin{figure}[th]
    \centering
    \includegraphics[width=0.9\columnwidth]{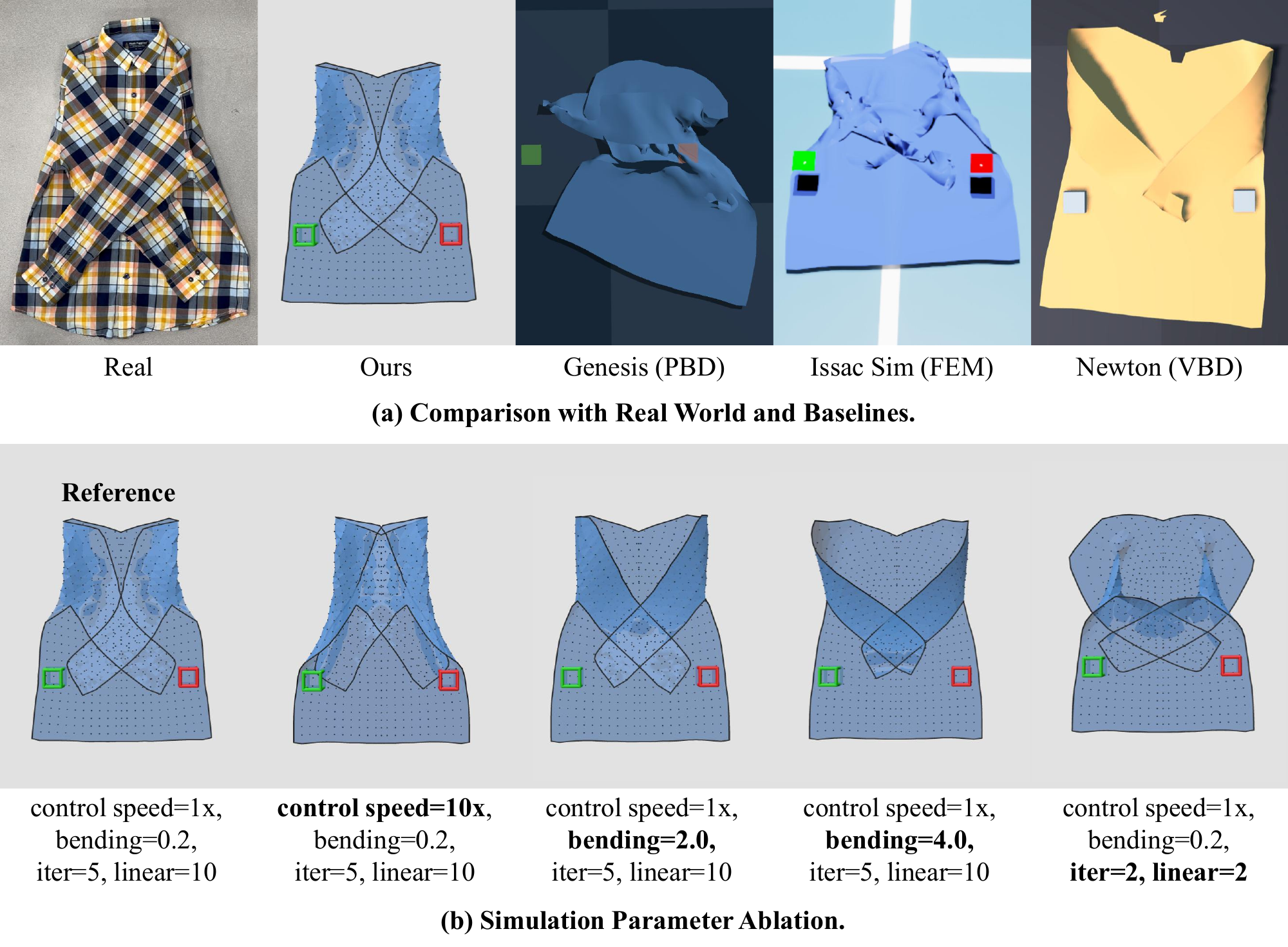}
    \caption{\textbf{T-shirt dual-sleeve folding results.} We compare the final results of different simulators and parameters after folding a T-shirt with a fixed trajectory. \textbf{(a) Comparison with real world and baselines.} We compare against representative GPU-capable simulators: Genesis (PBD), Isaac Sim (FEM), Newton (VBD), and real execution; FLASH most closely matches real behavior, producing smooth, symmetric folds that settle stably with accurate frictional sticking, while baselines exhibit various failure features. \textbf{(b) Simulation parameter ablation (FLASH).} With the trajectory fixed (Sec. V.A), varying control speed, bending stiffness, and solver iterations yields predictable changes. }
    \label{Fig: baseline}
    \vspace{-0.6\baselineskip}
\end{figure}
\begin{figure*}[t!]
    \centering
    \includegraphics[width=0.9\linewidth]{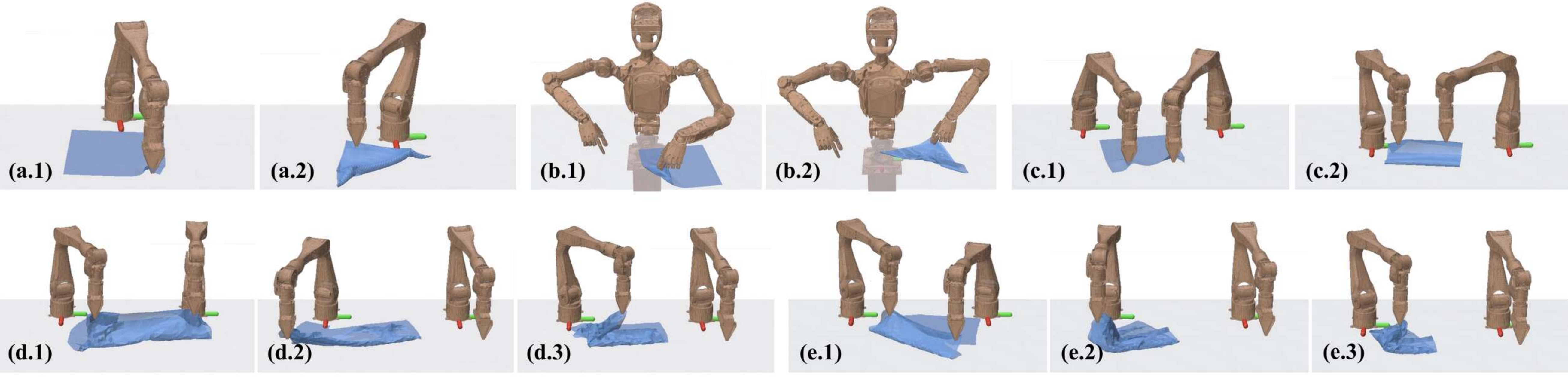}
    \caption{{\bfseries Learned policies in simulation.} (a) A single arm folds the towel. (b) A humanoid folds the towel. (c) Two arms fold the towel together. (d) Two arms fold the T-shirt. (e) Two arms fold the shorts.}
    \label{fig:learning_results_sim}
    \vspace{-0.3cm}
\end{figure*}
We compare FLASH against representative GPU-capable simulators: Isaac Sim (FEM), Newton (VBD), and Genesis (PBD), as well as real-world execution, with baseline parameters carefully tuned to maximize inextensibility, allow compliant bending, and avoid solver under-convergence.
Figure~\ref{Fig: baseline} shows that FLASH most closely matches real behavior, producing smooth, symmetric folds that settle into a stable resting configuration with accurate frictional sticking and no spurious sliding or post-release motion.
In contrast, Genesis exhibits large elastic snapping and poor frictional stability, leading to persistent sliding and failure to reach a stable fold.
Isaac Sim struggles to couple stiff in-plane stretching with compliant bending, resulting in shear-like distortions and excessive wrinkling reminiscent of crumpling.
Newton yields the strongest baseline result but remains overly stiff in bending: the folded sleeve fails to settle flush on the table and gradually elastically unfolds after release.
These results highlight that realistic garment folding critically depends on simultaneously accurate frictional contact resolution and well-balanced stretch–bend coupling.
Even small modeling or numerical artifacts can compound into qualitatively incorrect folds or unstable rest states, underscoring that high-fidelity contact-rich simulation is essential—not merely for visual realism, but for generating physically meaningful trajectories and states suitable for learning reliable manipulation policies.

\subsection{Parameter ablation in FLASH}
\label{Subsec:ablation}

We evaluate the physical interpretability of FLASH under variations in motion speed, bending stiffness, and solver fidelity (Fig.~\ref{Fig: baseline}(b)).
With slow manipulation and soft bending, FLASH produces smooth, symmetric folds that drape and settle stably.
Increasing the gripper speed by $10\times$ induces stronger inertial coupling, drawing the shoulder region tighter toward the fold as expected in dynamic pulling.
At fixed slow speed, increasing bending stiffness progressively lifts and widens the shoulder and increases elastic rebound, moving the sleeve cuff farther from the release point; at very high stiffness, the fold approaches over-flipping but remains stable due to gravity-elastic balance.
We finally push the solver to an extreme low-iteration regime by setting both the local-global iterations and the linear-solver iterations to 2. This substantially weakens long-range elastic coupling, so the grippers’ local motions can no longer fold the shoulder region effectively, however, the simulation still remains numerically stable and maintains a robust frictional solution.
Overall, FLASH parameters exhibit clear physical semantics, and reduced accuracy degrades behavior in predictable, robotics-relevant ways rather than producing unstable artifacts.


\subsection{Massively parallel throughput across simulators}
\label{Subsec: parallel}

We compare simulation throughput under large-scale parallel execution by measuring per-step time as the number of parallel environments increases (Table~\ref{tab:vb_throughput}), using solver settings that yield quasi-static folding behavior consistent with real execution rather than maximizing raw FPS.
Genesis becomes numerically unstable and diverges under multi-environment execution, preventing meaningful scaling.
Isaac Sim achieves strong parallel efficiency, but its simulated garment behavior remains physically inaccurate with pronounced artifacts (Figure~\ref{Fig: baseline}), limiting the usefulness of the generated data for learning.
Newton sustains stable execution but is consistently slower than FLASH and produces less realistic folding, whereas FLASH maintains stable simulation and offers a better fidelity/throughput tradeoff for the garment folding task.

\begin{table}[t]
    \centering
    \small
    \caption{
    \textbf{Massively parallel throughput comparison (ms/step; lower is better).} We vary the number of parallel environments from 1 to 256.
    }
    \label{tab:vb_throughput}
    \setlength{\tabcolsep}{4pt}
    \begin{threeparttable}
    \begin{tabular}{lcccccc}
        \toprule
        \textbf{Simulator} & \#1 & \#8 & \#32 & \#64 & \#128 & \#256 \\
        \midrule
        FLASH (ours) & 5.68 & 9.67 & 24.79 & 50.43 & 90.41 & 185.10 \\
        Genesis (PBD)& 5.97 & 6.07 & N/A\tnote{a} & N/A\tnote{a} & N/A\tnote{a} & N/A\tnote{a} \\
        Isaac Sim (FEM)  & 18.02 & 18.46 & 19.88 & 22.62 & 29.04 & 47.81 \\
        Newton (VBD) & 9.57 & 12.23 & 24.47 & 52.79 & 143.76 & 480.64 \\
        \bottomrule
    \end{tabular}
    \begin{tablenotes}[flushleft]
        \footnotesize
        \item[a] Simulation exhibits numerical instability and fails to converge.
        \vspace{-5mm}
    \end{tablenotes}
    \end{threeparttable}
\end{table}

\subsection{Learning Results in Simulation}

To move beyond simple pre-defined tasks that can be solved with hand-crafted controllers, we target more general real-world settings with diverse task specifications and initial conditions, and rely on learning-based policies.
As shown in Fig.~\ref{fig:learning_results_sim}, we use the proposed learning pipeline to train five cloth-folding tasks in simulation: single-arm towel folding with a desktop manipulator (a) and a humanoid robot (b), dual-arm towel folding (c), T-shirt folding (d), and shorts folding (e). The total wall-clock time (simulation plus learning) required to obtain deployable real-world policies on a single NVIDIA RTX 5090 is respectively 50 min for (a), 50 min for (b), 150 min for (c), 600 min for (d), and 600 min for (e).
These policies are then validated in the following section.

%% file: sections/6_real_robot.tex
\section{Real-Robot Experiments}~\label{Sec: Real-robot}
This section validates our approach on real robots. We begin by benchmarking efficiency and robustness on towel folding in Sec.~\ref{exp:real_towel}, followed by a demonstration of handling complex garments in Sec.~\ref{exp:complex_garment}, highlighting the framework's scalability to difficult manipulation scenarios.


\textbf{Experimental Setup.} 
We evaluate our method on two platforms shown in right part of Fig.~\ref{Fig: teaser}: \textbf{Airbot Play}~\cite{airbot_web} (a pair of 6-DoF desktop arms with a parallel gripper) and \textbf{AdamU}~\cite{adamu_web} (an upper-body humanoid with dual arms and dexterous hands). 
Both utilize a top-down ZED Mini for depth stream perception. 
Inference and control are handled by an external RTX 4090 workstation for Airbot, or an onboard Jetson Orin NX and Intel NUC for AdamU.
The policy predicts incremental end-effector positions and gripper open/close commands. 
These are executed via the manufacturer's SDK (Airbot) or processed through a numerical IK solver~\cite{Zhong_PyTorch_Kinematics_2024} with low-level PID tracking (AdamU).


\subsection{Folding Towel}
\label{exp:real_towel}
\subsubsection{Efficiency and Robustness Analysis}
Training on a single NVIDIA RTX 5090, our framework yields a robust policy on the desktop Airbot Play manipulator in 50 minutes. Despite this rapid training, the closed-loop policy demonstrates:

\textbf{Robust Initialization.} 
The system tolerates significant placement uncertainty, succeeding under spatial translations of $\pm 8$\,cm which effectively spans the reachable workspace and arbitrary azimuthal rotations.

\begin{figure}[t]
    \centering
    \includegraphics[width=0.95\columnwidth]{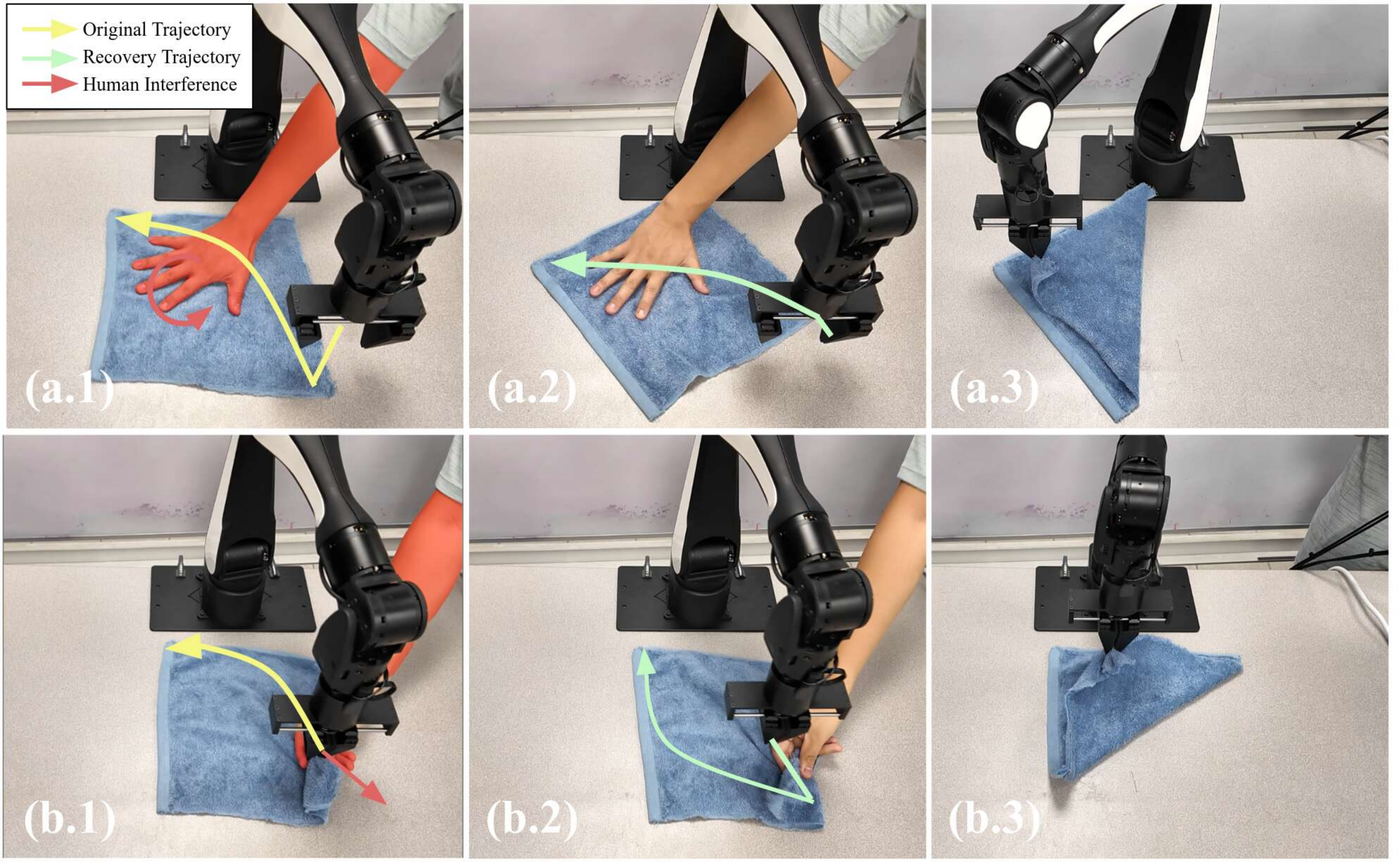}
    \vspace{-5pt}
    \caption{\textbf{Robustness to external disturbances.} (a.1–a.3) Recovery from unexpected \textbf{displacement} by human. (b.1–b.3) Recovery from the towel being \textbf{pulled away}. The system maintains task continuity without human intervention.}
    \label{Fig: recovery}
    \vspace{-10pt}
\end{figure}

\textbf{Reactive Recovery.} 
As shown in Fig.~\ref{Fig: recovery}, the policy exhibits emergent recovery behaviors against dynamic disturbances:
\begin{itemize}
    \item \textit{Missed Grasps:} Driven by continuous visual feedback, the robot naturally re-attempts grasp actions upon failure.
    \item \textit{Human Interference:} The policy dynamically adapts to external perturbations (e.g., dragging the towel away), ensuring task completion without reset.
\end{itemize}




\subsubsection{Long-Duration Continuous Evaluation}

\begin{figure}
    \centering
    \includegraphics[width=0.9\linewidth]{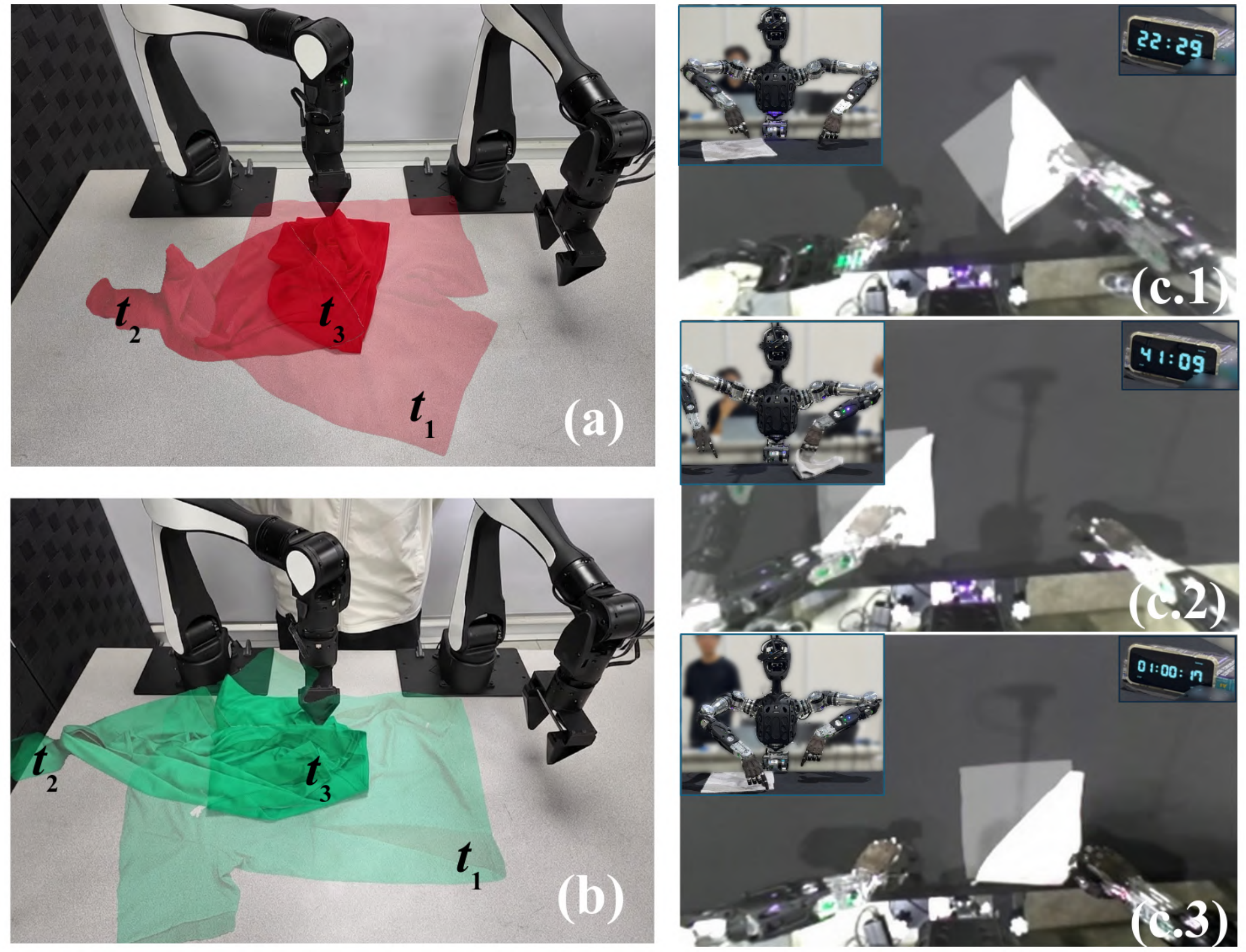}
    \caption{\textbf{Qualitative real-world results.} 
    (Left) Keyframe visualization of bimanual folding ($t_1 \to t_3$) for (a) shorts and (b) T-shirt. 
    (Right) Snapshots from the continuous evaluation on AdamU at 20, 40, and 60\,min (c.1--c.3). 
    Main frames show egocentric camera observations; insets show the third-person view and clock time.}
    \label{fig:real_qualitative_results}
    \vspace{-0.5cm}
\end{figure}

We conducted a continuous one-hour evaluation on AdamU, utilizing both arms to assess system consistency. 
Over the course of 106 back-to-back trials, a trial is considered successful if the robot aligns the towel corner within 5\,cm of the goal pose in under 40 seconds. 
Under this metric, the system achieved an aggregate 85.8\% success rate (91/106). 
Figure~\ref{fig:real_qualitative_results} visualizes these qualitative results by overlaying the transparent initial states with the opaque final states. 
This visualization confirms that despite significant initialization variance and accumulated cloth deformations, the policy consistently converges to the precise target configuration. 
Furthermore, inference latency remained stable throughout the extended operation.


\subsection{Folding Tasks on Other Garments}
\label{exp:complex_garment}
    
To validate the generalization capability of our framework, we extend the experiments to garments with distinct topologies and dynamics. Here, we focus on the \textit{consistency} between simulation behaviors and real-world execution, shown in Fig.~\ref{fig:real_qualitative_results}.

\subsubsection{Folding Shorts}
This experiment involves a folding task for short pants with a dual-leg topology. The robot first aligns and folds the two pant legs, followed by a lateral fold from one side. Qualitative results demonstrate a successful sim-to-real transfer, with the policy effectively handling the geometric split and physical complexity. Quantitatively, our system achieved a 60\% success rate (12/20) under human evaluation. 

\subsubsection{Folding T-shirts}
Folding T-shirts constitutes a longer-horizon task compared to single-step folding and introduces substantial variation during execution, making it particularly challenging. We demonstrate that the learned manipulation policy successfully performs the required multi-step folding sequence in the real world, closely matching its behavior in simulation.
In experiments, the system achieved consecutive successful executions, indicating that our sim-to-real pipeline preserves trajectory coherence and physical plausibility even over long-horizon manipulation sequences. Quantitatively, our system achieved a 70\% success rate (35/50) under human evaluation. 

\subsection{Failure Modes}

We identify two dominant failure sources in deployment:
\subsubsection{Perception bottleneck} We use segmented depth inputs to isolate dynamics transfer from the RGB visual gap. Depth sensor noise on thin fabrics and self occlusions cause occasional grasp misalignment, accounting for the majority of real-world failures and the imprecise final geometry.
\subsubsection{Hardware abstraction gap} We abstract diverse hardware
(AdamU, Airbot) into a unified binary grasp model without
modeling motor level dynamics such as actuation delays
and backlash. This enables zero-shot transfer but introduces
tracking deviations that degrade final garment geometry. The lack of tactile feedback further limits the system’s ability to detect and correct these deviations.



%% file: sections/7_conclusion.tex
\section{Conclusion and Discussion}~\label{Sec: Conclusion}
In this paper, we presented FLASH, a GPU-native simulation platform for deformable simulation, rendering, and learning. By coupling GPU-parallel simulation with efficient depth rendering and a scalable training pipeline, FLASH supports high-throughput data generation for deformable manipulation learning. Through a suite of garment manipulation tasks, we demonstrate that FLASH reduces the sim-to-real gap and enables simulation-only training across multiple contact-rich scenarios. Notably, our vision-based policies can be trained in minutes and transferred zero-shot to real hardware, exhibiting robust execution and recovery behaviors. Our results strengthen the case for using large-scale synthetic interaction data to train deformable manipulation policies, substantially reducing reliance on costly real-world data collection. However, limitations remain. Our current implementation still incurs CPU–GPU transfer overhead in parts of the pipeline, leaving room for further optimization. In addition, handling more complex and longer-horizon garment manipulation will likely require improved learning formulations and supervision signals, including more effective reward or feedback design.

%% file: sections/10_acknowledgement.tex
\section*{Acknowledgment}

This work was supported in part by the NUS Presidential Young Professorship grant (A-0009982-00-00) and the MOE AcRF Tier 1 24-1234-P0001 Funding. The authors would like to thank the Gradient and PranaLabs for their generous support and for providing the collaborative environment necessary to conduct this research. We also thanks the support from Swiss AI Initiative Small Grant, NVIDIA Academic Grant, and Google Research Funding.

%% file: sections/appendix.tex

\section*{Supplementary Materials}

\begin{figure*}[b!]
    \centering
    \includegraphics[width=\linewidth]{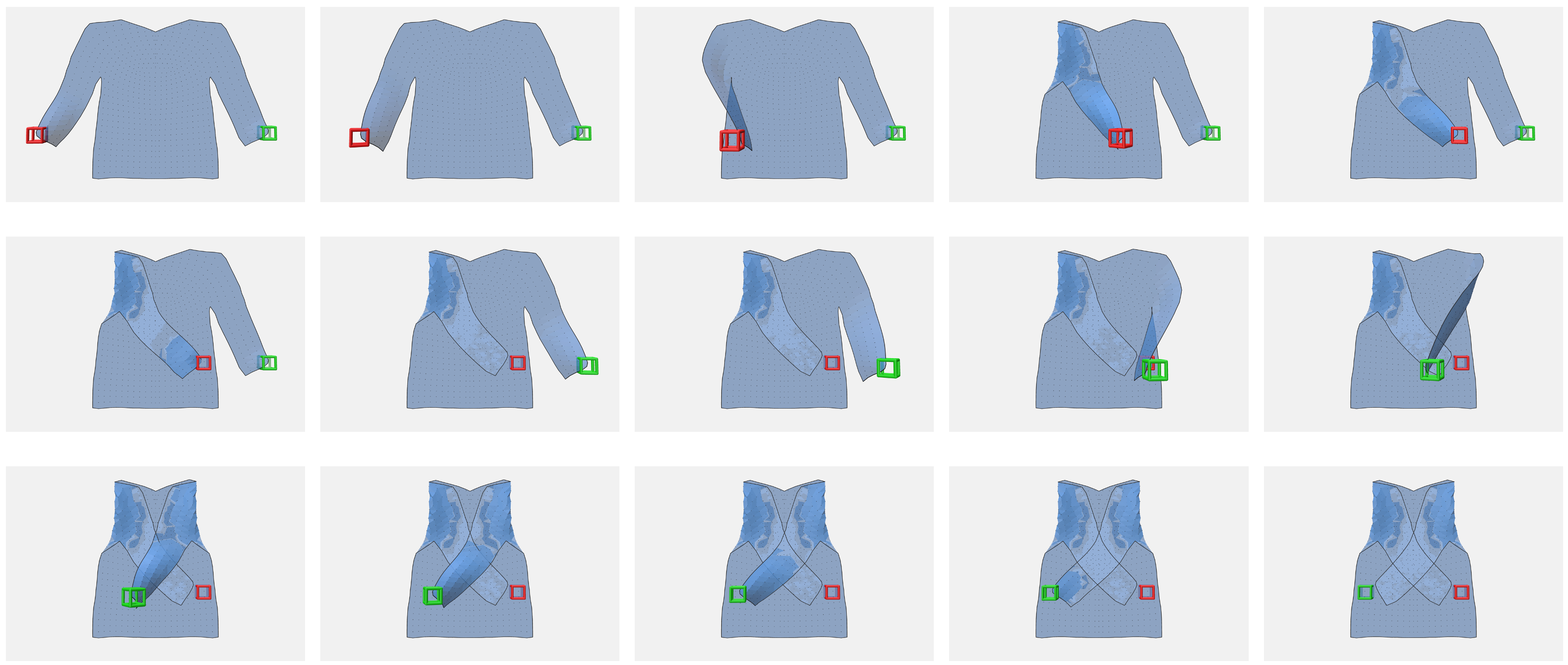}
    \caption{{\bfseries Snapshots from the bimanual shirt folding example, simulated using our FLASH simulator.} The dual end-effectors execute a sequential folding strategy: the left gripper first manipulates the left sleeve to the center and releases it, followed by the right gripper completing the fold for the opposing sleeve. The end-effector trajectories are generated via kinematic interpolation through a sequence of pre-defined heuristic waypoints.}
    \label{fig:shirt_fold_steps_appendix}
\end{figure*}

\begin{figure*}[tbh!]
    \centering
    \begin{minipage}{0.49\textwidth}
        \begin{lstlisting}[
            basicstyle=\ttfamily\scriptsize, % Smaller font for JSON to fit
            frame=single, 
            title={\textbf{(a) Scene Configuration (scene.json)}}
        ]
{
    "timestep": 0.01,
    "gravity": [0, -9.81, 0],
    "linearsolver": { "type": "SPARSE_INVERSE_CUDA" },
    "constraintsolver": {
        "type": "LiteNonSmoothNewton_CUDA",
        "maxforce": 1E+10,
        "iterations": 10,
        "tolerance": 1E-9
    },
    "planecollisions": { ... },
    "objects": {
        "cloth": "path/to/cloth.json"
    }
}
        \end{lstlisting}
    \end{minipage}
    \hfill
    \begin{minipage}{0.49\textwidth}
        \begin{lstlisting}[
            basicstyle=\ttfamily\scriptsize, 
            frame=single, 
            title={\textbf{(b) Asset Properties (cloth.json)}}
        ]
{
    "mesh": "path/to/cloth.obj",
    "element_type": "TRIANGLE",
    "transformation": {
        "trans": [0, 0, 0],
        "rotation": [0, 0, 0]
    },
    "mechanical_props": {
        "obj_mass": 1000,
        "young": 3E+4,
        "poisson": 0.4,
        "constitutive": "TRI_ARAP",
        "bending": 0.2
    }
}
        \end{lstlisting}
    \end{minipage}

    \vspace{2pt}
    
    \begin{lstlisting}[
        language=Python, 
        basicstyle=\ttfamily\small, 
        frame=single, 
        title={\textbf{(c) Control Script (Python)}}
    ]
import flash_sim

# 1. Initialize Simulator
sim = flash_sim.RealtimeSim()
sim.load_config("path/to/scene.json") # Load config defined in (a)
sim.set_envs(n_envs=128)              # Activate parallel environments
sim.initialize()

# 2. Setup Dual End-Effectors
sim.add_ee("left",  start_pose_l, size=[0.05]*3)
sim.add_ee("right", start_pose_r, size=[0.05]*3)

# 3. Simulation Loop
for step in range(1000):
    # Trajectory generation
    pose_l, grasp_l = compute_fold_trajectory(step, side="left")
    pose_r, grasp_r = compute_fold_trajectory(step, side="right")

    sim.move_ee("left",  pose_l)
    sim.grasp_ee("left", grasp_l)
    sim.move_ee("right", pose_r)
    sim.grasp_ee("right", grasp_r)

    sim.step()
    \end{lstlisting}

    \vspace{-5pt}
    \caption{\textbf{Implementation overview for the bimanual shirt folding example (Fig. \ref{fig:shirt_fold_steps_appendix}).} 
    \textbf{(a)} The scene configuration defines GPU-accelerated solvers and environment settings. 
    \textbf{(b)} The asset configuration defines the object's initial transformation and physical properties (e.g., bending stiffness).
    \textbf{(c)} The high-level Python API allows users to drive massively parallel simulations (e.g., $128$) with concise control logic. Note that auxiliary code (e.g., imports, trajectory interpolation) is omitted for clarity.}
    \label{fig:code_and_config}
\end{figure*}

\section{Simulation System Specification}
To demonstrate the usability and modularity of our simulation framework, we provide the complete implementation details for the bimanual shirt folding example presented in Fig. \ref{fig:shirt_fold_steps_appendix}.
Our system adopts a hybrid architecture: the core physics engine is implemented in highly optimized C++ and CUDA to ensure maximum efficiency on the GPU, while the user interface is exposed via Python bindings (pybind11) to facilitate rapid prototyping and ease of use.

As shown in Fig. \ref{fig:code_and_config}, the simulation workflow is structured into two logical components:

\subsubsection{Simulation Configuration}
We configure the underlying physics engine through two distinct specifications: one establishes global simulation settings (e.g., gravity, timestep), static boundaries (e.g., plane collisions), and numerical solvers, while the other defines object-specific attributes (e.g., initial transformation and mechanical properties).
\begin{itemize}
    \item \textbf{Scene Configuration (Fig. \ref{fig:code_and_config} (a)):} This file establishes the global numerical environment and static boundaries. It defines fundamental simulation constants such as \texttt{timestep} and \texttt{gravity}, alongside environmental primitives like \texttt{planecollisions}. Crucially, it configures the solver, specifically the \texttt{LiteNonSmoothNewton\_CUDA} constraint solver, while exposing convergence parameters (e.g., \texttt{iterations}, \texttt{tolerance}) to control the precision of the solver.
    
    \item \textbf{Asset Properties (Fig. \ref{fig:code_and_config} (b)):} This file defines the object's geometric initialization and constitutive behavior. It specifies the initial kinematic state via the \texttt{transformation} field and selects the discretization element type (e.g., \texttt{TRIANGLE} and \texttt{TETRAHEDRON}). Mechanical fidelity is strictly governed by the \texttt{mechanical\_props} part, where we specify the elastic energy type (e.g. \texttt{TRI\_ARAP} for As-Rigid-As-Possible energy) and parameters such as \texttt{obj\_mass} and \texttt{bending} stiffness are explicitly assigned to determine deformation characteristics.
\end{itemize}

\subsubsection{High-Level Python Interface}
Fig. \ref{fig:code_and_config} (c) demonstrates how to use the Python API to drive the simulation. The workflow proceeds as follows:
\begin{itemize}
    \item \textbf{Initialization:} The script first loads the scene configuration using \texttt{sim.load\_config()} and sets the parallel environment count via \texttt{sim.set\_envs()}. The C++ backend parses the JSON files and automatically initializes the specified number of parallel environments, allocating necessary GPU memory.
    \item \textbf{Task Setup:} We add manipulators using high-level commands like \texttt{sim.add\_ee()}. Additionally, we also provide rendering-related API like setting up cameras.
    \item \textbf{Control Loop:} In the main loop, the system takes as input kinematic targets (e.g., via policy output) and sends commands via \texttt{sim.move\_ee()} and \texttt{sim.grasp\_ee()}. The \texttt{sim.step()} function then invokes the physics engine of FLASH to advance the simulation state for all parallel environments simultaneously.
\end{itemize}

\FloatBarrier
\vspace{1cm}
\section{Real-to-Sim Details}
To construct a high-fidelity simulation environment that minimizes the reality gap, we address two critical aspects: geometric consistency and dynamic fidelity. We first detail the \textbf{extrinsic alignment} of the robot workcell to ensure spatial correspondence. Subsequently, we describe the \textbf{system identification} process used to tune the physical parameters of the deformable object.
\begin{figure}[htb]
    \centering
    \includegraphics[width=0.6\linewidth]{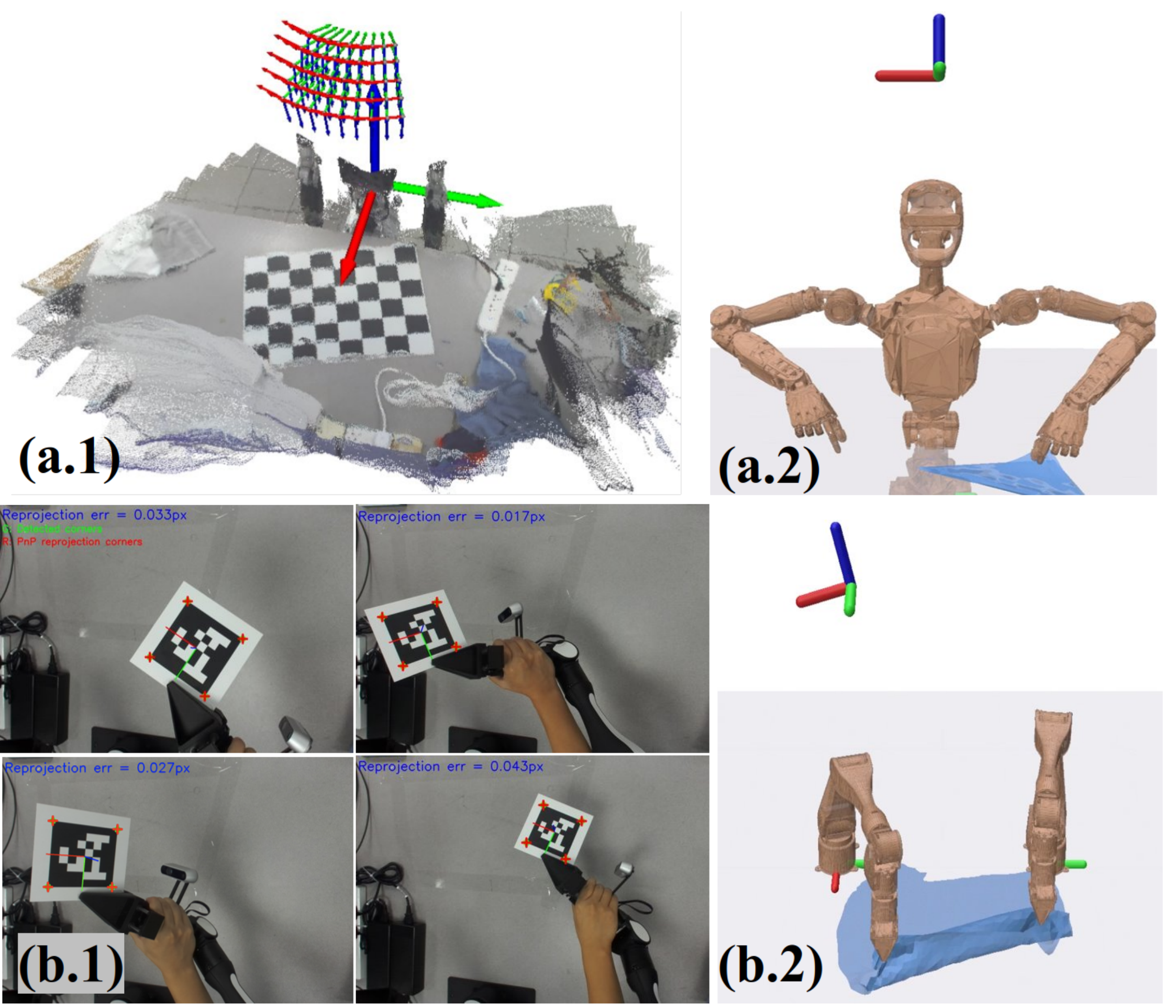}
    \caption{\textbf{Visualization of Extrinsic Alignment and Scene Calibration.} 
    \textbf{(a.1) \& (b.1) Real-world Validation:} The accumulated point clouds for Adam-U (a.1) and the sub-pixel reprojection errors ($<0.05$\,px) for Airbot (b.1) confirm high-precision geometric alignment.
    \textbf{(a.2) \& (b.2) Simulation Alignment:} The corresponding digital twin environments are shown on the right. In both setups, the calibrated camera frame is visualized overhead, where the \textbf{blue arrow} explicitly denotes the Z-axis aligned with the optical axis.}
    \label{fig:scene_calib}
\end{figure}
\subsubsection{Extrinsic Alignment}
We present the specific calibration procedures and validation results for our two experimental setups below.

\paragraph{AdamU Setup (Eye-in-Hand)} 
To achieve a comprehensive top-down view, we mounted a ZED Mini camera via a custom rigid extension above the robot's head. Since the camera moves with the neck joints (yaw and pitch), we formulated the alignment as an eye-in-hand calibration problem. We collected approximately 40 RGB-D frames of a static checkerboard from diverse viewpoints and solved for the kinematic transformation. Validation via dense point cloud accumulation (Fig.~\ref{fig:scene_calib}) demonstrates high consistency; the superimposed checkerboard pattern remains sharp and planar across all views, confirming precise alignment. Note that minor geometric artifacts at the periphery are attributable to the intrinsic depth distortion of the ZED Mini sensor at oblique angles rather than calibration error.

\paragraph{Airbot Setup (Eye-to-Hand)} 
For the Airbot platform, we employ an eye-to-hand configuration with a fixed external top-down camera. The robot's end-effector grasps a lightweight (100g) calibration board featuring a single 9\,cm ArUco marker. Following a standard eye-to-hand calibration pipeline, we collect a series of RGB images paired with end-effector poses retrieved directly from the Airbot API. We validate the calibration accuracy by comparing the marker's geometric center and corners, computed via the robot's forward kinematics and the solved extrinsics, against their ground-truth pixel coordinates detected in the image. As illustrated in Figure~\ref{fig:scene_calib}, the projected corners align closely with the visually detected ones, demonstrating minimal reprojection error and precise spatial registration.
For our dual-arm configuration, this procedure is performed independently for each manipulator. By registering both robot bases relative to the shared fixed camera frame, we effectively unify the entire workspace into a single global coordinate system, enabling precise bi-manual coordination.

\subsubsection{System Identification}
\label{sec:sysid}

We aim to identify the physical parameters of the deformable object that minimize the behavioral discrepancy between simulation and reality. We focus on tuning the material properties governing deformation, specifically Young's modulus and Poisson's ratio.

\paragraph{Data Collection}
We design a canonical "corner lift" interaction to fully excite the material's dynamic properties. The robot grasps one corner of the fabric and executes a lifting trajectory to induce complex deformations. To capture the ground truth geometry, we record the sequence using a depth camera. We segment the cloth from the background to extract the dynamic point cloud sequence. Note that parts of the real-world cloth point cloud may be incomplete due to occlusions caused by the robot gripper during the lift.

\begin{figure}[htb]
    \centering
    \includegraphics[width=0.8\linewidth]{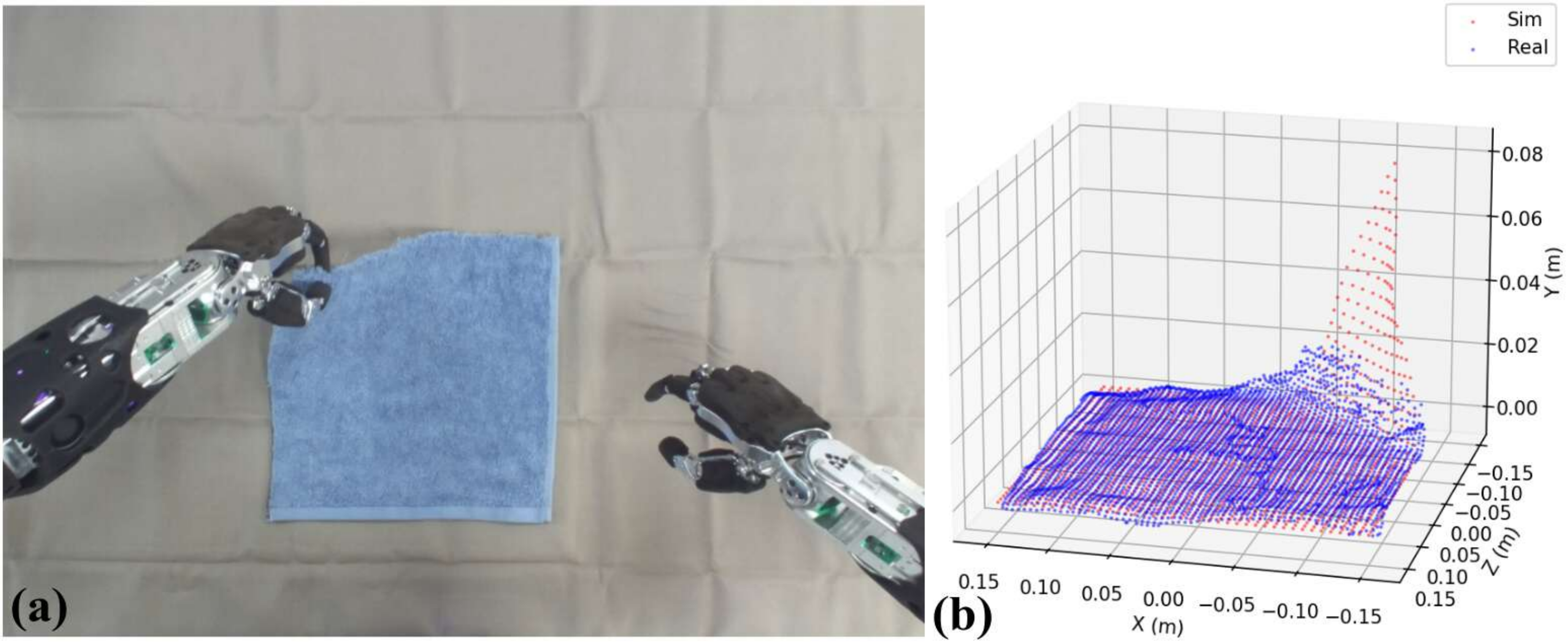}
    \caption{\textbf{Visualization of System Identification.} (a) Real-world experimental setup for data collection. (b) Spatial alignment between the simulated mesh (red) and real-world observation (blue) at a synchronized timestamp. Note that the missing upper region of the real-world point cloud is due to occlusion by the robot gripper.}
    \label{fig:sys_id}
\end{figure}

\paragraph{Parameter Optimization}
We employ an open-loop replay strategy to reproduce the experiment in simulation. Specifically, we drive the simulated robot using the exact end-effector trajectory recorded from the real world. We perform a grid search over Young's modulus and Poisson's ratio (while keeping density and friction fixed based on physical measurements). 

For each parameter set, we generate a simulated mesh sequence. The optimal parameters are selected by minimizing the geometric alignment error between the real and simulated states. We quantify this error using the average point-to-mesh distance between the segmented real-world point clouds and the simulated surface at synchronized keyframes. As visualized in Figure~\ref{fig:sys_id}, the simulated mesh (red) exhibits strong spatial alignment with the observed real-world point cloud (blue) under the identified parameters, illustrating the geometric correspondence achieved by our dynamic modeling.

\begin{figure}[ht] 
    \centering
    \includegraphics[width=0.6\linewidth]{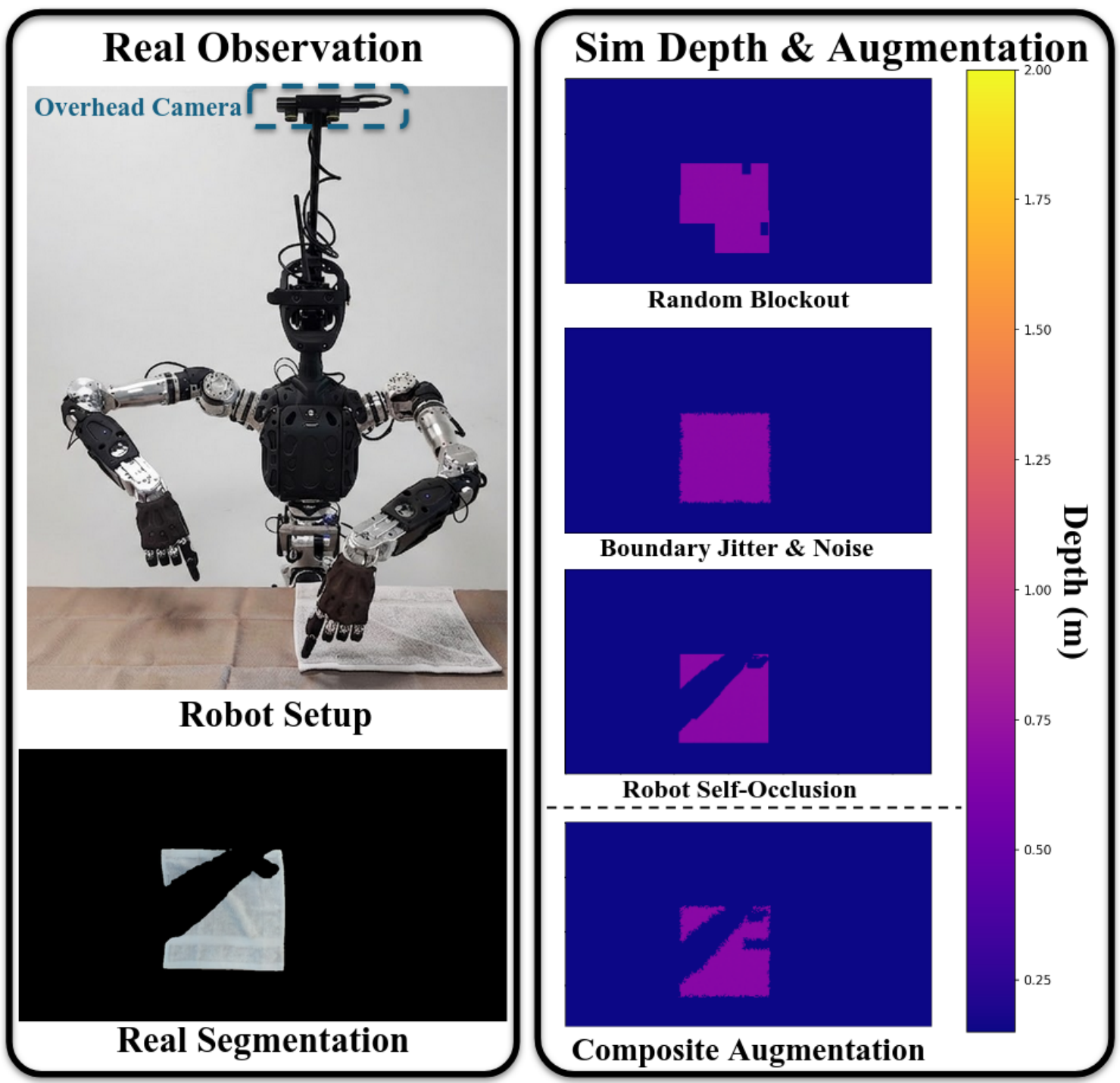}
    
    \caption{\textbf{Real-Sim Observation Alignment and Depth Augmentation.} 
    \textbf{Left:} The physical hardware setup featuring an overhead RGB-D camera and the resulting \textit{segmented RGB observation}. Note that the real observation (bottom left) presents obvious robot arm occlusions.
    \textbf{Right:} Our simulation rendering pipeline designed to bridge this domain gap. We sequentially apply \textit{random blockout} to mimic segmentation imperfections, \textit{boundary jitter \& noise} to match sensor characteristics, and \textit{robot self-occlusion} to reproduce the physical setup. The final \textit{composite augmentation} (bottom right) serves as the robust input for policy training.}
    
    \label{fig:sim2real_perception}
\end{figure}

\subsubsection{Depth Simulation and Perception Augmentation}
\label{sec:depth_sim}

To ensure the policy transfers zero-shot to the real world, the perception pipeline in simulation must mimic both the sensor-level imperfections and the semantic-level segmentation errors observed in real-world experiments.

Figure~\ref{fig:sim2real_perception} visualizes the \textit{alignment of depth observations} between the physical setup and the simulation environment. 
The left side depicts the hardware setup with an overhead RGB-D camera. 
We visualize the \textit{segmented RGB observation} (bottom left) to intuitively display the robot's configuration and occlusion challenges; however, strictly speaking, our policy operates solely on the corresponding \textit{segmented depth images}.
As shown in the comparison, the raw real-world data presents obvious artifacts, including robot self-occlusion and imperfect segmentation boundaries. 
To bridge this gap, our simulation pipeline (right) systematically introduces synthetic perturbations---including random blockout, boundary jitter, and kinematic-based self-occlusion---generating a \textit{composite augmented observation} that closely matches the real-world distribution.

In addition to mitigating the depth rendering gap between simulation and the real world, it is also crucial to maintain high computational efficiency to facilitate massively parallel reinforcement learning. 
We implement a versatile ray-casting pipeline that generates depth maps, instance segmentation masks, and point clouds directly from the simulation state. 
To accommodate diverse hardware infrastructures, our system supports both a GPU-based backend utilizing the NVIDIA OptiX and a CPU-based backend powered by Intel Embree. 
As shown in Table~\ref{tab:time_depth_render}, we evaluate the rendering overhead across increasing environment counts. 
While the current rendering time scales linearly with the number of environments due to the present engineering implementation of the ray-casting batches, the absolute latency remains sufficiently low to meet the high-throughput requirements of policy learning. 
This linear trend is primarily driven by the overhead of dynamic vertex update, where future implementations utilizing batched geometry updates will further reduce rendering latency.

\begin{table}[tbh!]
    \centering
    \small
    \caption{\textbf{Depth rendering overload over massively parallel environments.} We evaluate on the AdamU towel-folding scene ($32\mathrm{k}$ vertices, $13\mathrm{k}$ faces).}
    \label{tab:time_depth_render}
    \setlength{\tabcolsep}{4pt}
    \begin{tabular}{lcccccc}
        \toprule
        \textbf{Num. Envs} & 1 & 8 & 32 & 64 & 128 \\
        \midrule
        \textbf{Time (ms)} & 0.75 & 2.23 & 7.20 & 14.18 & 27.31 \\
        \bottomrule
    \end{tabular}
\end{table}

\FloatBarrier
\vspace{1cm}
\section{Teacher-Student Learning Designs}

\subsubsection{Teacher Synthesis}

\begin{figure*}[!htb]
    \centering
    \begin{tikzpicture}[
        font=\small\sffamily, 
        >=Stealth, 
        line width=1pt
    ]
    \definecolor{myblue}{RGB}{0, 75, 160}
    \definecolor{myred}{RGB}{180, 30, 30}
    \definecolor{mygreen}{RGB}{0, 110, 50}

    \begin{scope}[local bounding box=traj]
        \coordinate (Start)    at (0,    1.6);   
        \coordinate (HoverEnd) at (1.8,  0.6);   
        \coordinate (Keypoint) at (2.4, -0.9);   
        \coordinate (Goal)     at (6.0,  0.3);     
           
        \node[anchor=west, xshift=-20pt, yshift=0pt] at (Keypoint)
          {\includegraphics[width=5.cm]{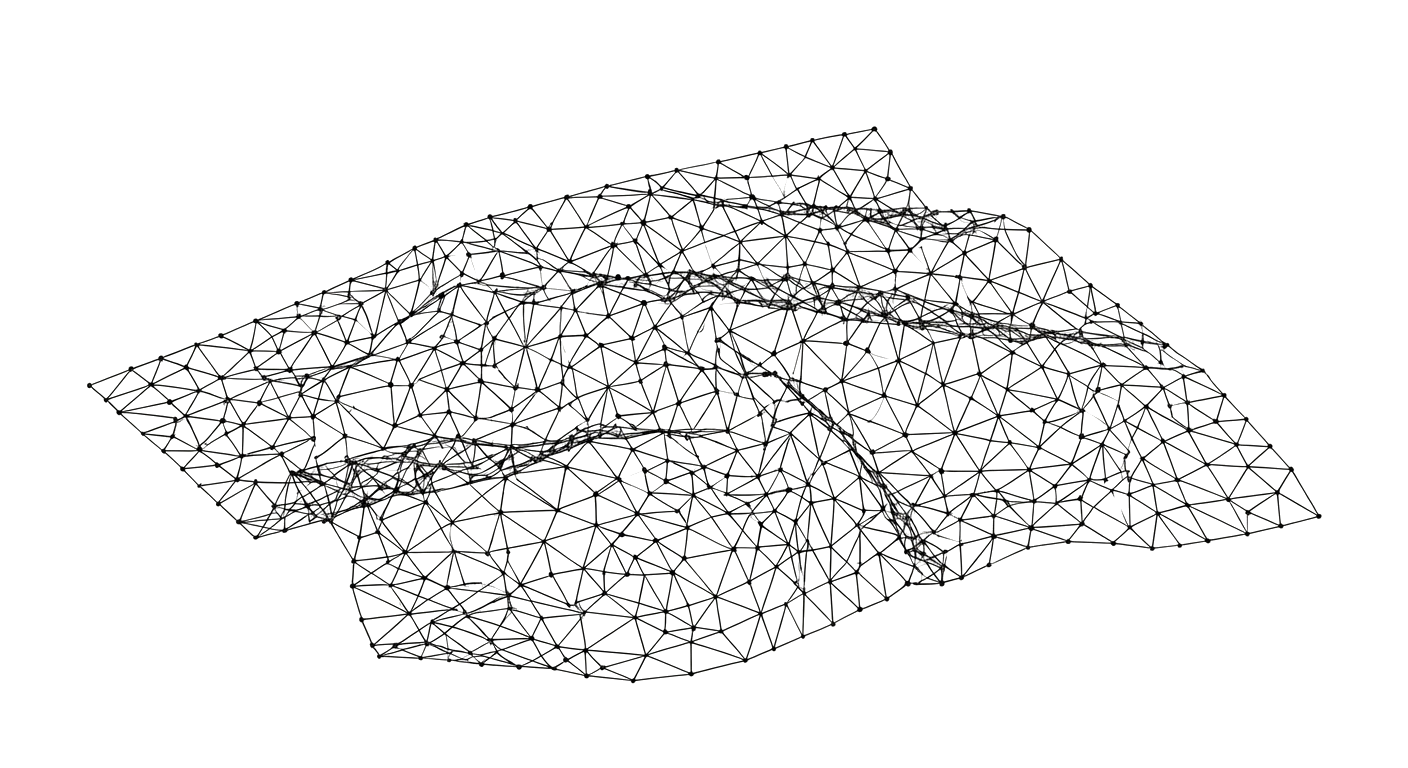}};
            
        \draw[->, myblue, dashed, line width=1.2pt] (Start) to[out=0, in=160] (HoverEnd);
        \node[myblue, above, font=\bfseries, yshift=4pt] at (0.7, 1.6) {1. Approach};
    
        \draw[->, myred, line width=1.3pt] (HoverEnd) -- (Keypoint);
        \node[myred, font=\bfseries] at (1.3, 0.0) {2. Grasp};
    
        \draw[->, mygreen, line width=1.3pt] (Keypoint) to[out=70, in=170] (Goal);
        \node[mygreen, font=\bfseries] at (4.0, 0.7) {3. Transport};

        \fill[black] (Keypoint) circle (3pt) node[below=7pt, left=5pt, text=black] {Keypoint};
        \fill[black] (Goal) circle (3pt) node[above=3pt, text=black] {Target};
    \end{scope}

    \draw[gray!20, line width=2pt] 
        (7.4, -2.7) -- (7.4, 3.2);

    \begin{scope}[shift={(13.5, 0)}, local bounding box=fsm]
        \tikzset{state/.style={
            draw, rounded corners=3pt, 
            minimum width=4.5cm, minimum height=1.2cm, 
            align=center, line width=1pt, fill=white}
        }
        
        \node[state, draw=myred, fill=myred!5] (StGrasp) 
            {\textbf{2. GRASP}\\ \footnotesize(Reach down \& Close EE)};
        
        \node[state, draw=myblue, fill=myblue!5, above=1.0cm of StGrasp] (StApp) 
            {\textbf{1. APPROACH}\\ \footnotesize(EE Open)};

        \node[state, draw=mygreen, fill=mygreen!5, below=1.0cm of StGrasp] (StTrans) 
            {\textbf{3. TRANSPORT}\\ \footnotesize(EE Closed)};

        \draw[->] (StApp) -- (StGrasp) 
            node[midway, right, font=\scriptsize, align=left, xshift=8pt] {Horizontal Dist\\$<$ Threshold};
        
        \draw[->] (StGrasp) -- (StTrans) 
            node[midway, right, font=\scriptsize, align=left, xshift=8pt] {EE Closed\\\& Contact Detected};

        \draw[->, rounded corners=12pt] (StTrans.west) -- ++(-1.2, 0) 
            |- (StApp.west)
            node[pos=0.25, left, align=right, font=\scriptsize, xshift=-10pt] 
            {Object Dropped\\(Recovery)};
    \end{scope}

    
    \node[anchor=north, font=\bfseries] (labelA) at ($(traj.center) + (0,-3.5)$) 
        {(a) Illustration of teacher EE trajectory};
    
    \node[anchor=north, font=\bfseries] at (fsm.center |- labelA.north) 
        {(b) Low-level primitive state machine};

    \end{tikzpicture}
    \caption{\textbf{Low-level teacher primitives.} Our teachers use the depicted low-level primitives to grasp and transport specific keypoints to target locations.}
    \label{fig:primitives}
\end{figure*}

We generate teacher actions from cloth-state information using a hierarchical finite-state machine design. As illustrated in Fig.~\ref{fig:primitives}, we design a low-level pattern for the end-effector (EE) to grasp and transport keypoints via:
\begin{itemize}
    \item an \textit{Approach} primitive that maintains a vertical hover while moving laterally until the EE is horizontally aligned with the keypoint vertex,
    \item a \textit{Grasp} primitive that reaches down and closes the EE,
    \item and a \textit{Transport} primitive that moves the grasped keypoint toward the target.
\end{itemize}
When a grasp fails, the teacher goes back to the \textit{Approach} phase to recover.

The high-level sequencing of these primitives is adapted to each task's topology (Table~\ref{tab:high_level_logic}). For towel folding tasks, the teacher performs direct single-stage folds. For T-shirt and shorts folding, we use a multi-stage procedure that includes dragging to exploit the limited workspace and complete the fold. The corresponding motion patterns are also shown in our videos. We specify stage-dependent thresholds to detect failures and trigger a return to earlier stages when necessary.
\\

\subsubsection{Student Architecture}
\begin{figure*}[!htb]
    \centering
    \includegraphics[width=1\linewidth]{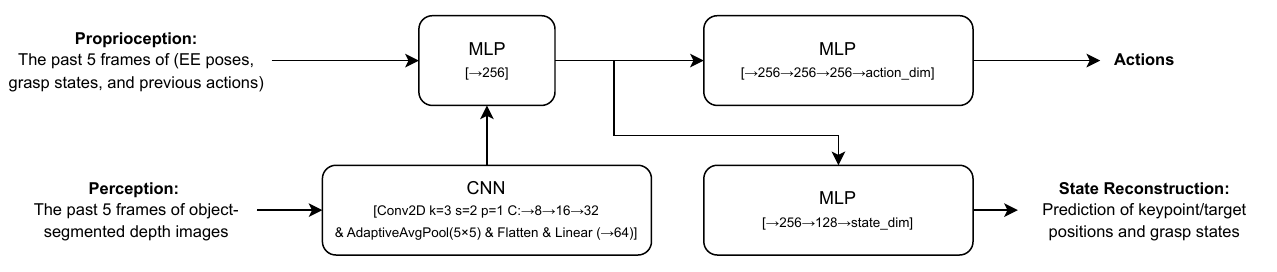}
    \caption{\textbf{Student policy architecture.} Our student policy outputs actions and state reconstruction from proprioception and perception observations.}
    \label{fig:student-arch}
\end{figure*}

Our student policy architecture is illustrated in Fig.~\ref{fig:student-arch}. We concatenate five-step histories of both proprioceptive and perceptual inputs. Perception is encoded with a convolutional neural network (CNN), while the remaining components are implemented with multilayer perceptrons (MLPs). We additionally train the model to reconstruct state variables, which helps automatic fulfillment and termination detection during deployment.
\\

\subsubsection{Training Details}

We apply DAgger~\cite{ross2011reduction} to distill teacher actions into deployable student policies, using the mean-absolute error (MAE) loss for position delta and the log-probability loss for EE open/close logits. In addition to the action distillation objectives, we include an auxiliary mean-squared error (MSE) loss for state reconstruction.

During training, we apply domain randomization to the cloth dynamics parameters ($0.5\times \sim 1.5\times$), the initial cloth and EE states, proprioceptive observations (uniform noise calibrated to empirical control precision), and perception (camera pose jitter, edge noise, and random occlusions).

The implementation will be open sourced for reproducibility.

\begin{table*}[!htb]
\centering
\begin{threeparttable}
\caption{High-Level Teacher Scheduling for Cloth Folding Tasks}
\label{tab:high_level_logic}
\begin{tabular}{l c c c c}
\toprule
\textbf{Task} & \textbf{Stage} & \textbf{EE} & \textbf{Keypoint} & \textbf{Target} \\
\midrule
Towel (1 arm) & 1 & Left & Left-top corner & Right-bottom corner \\
\midrule
Towel (2 arm) & 1 & Left & Distant left corner & Adjacent left corner \\
              &   & Right & Distant right corner & Adjacent right corner \\
\midrule
T-Shirt & 1 & Left & Left hem & Right hem \\
        &   & Right & Left sleeve & Right sleeve \\
        & 2 & Left & No grasp & Point in mid-air \\
        &   & Right & Neckline & Point on the right \\
        & 3 & Right & Neckline & Center hem \\
\midrule
Shorts  & 1 & Left & Left opening & Right opening \\
        &   & Right & Left waist & Right waist \\
        & 2 & Left & No grasp & Point in mid-air \\
        &   & Right & Center waist & Point on the front-right \\
        & 3 & Right & Center waist & Right opening \\
\bottomrule
\end{tabular}
\begin{tablenotes}
    \small
    \item[*] All designs can be mirrored for left- versus right-sided execution.
\end{tablenotes}
\end{threeparttable}
\end{table*}

\FloatBarrier
\vspace{1cm}
\section{Sim-to-Real Details}
\subsubsection{Online Segmentation Pipeline}
We adopt a streamlined combination of \textbf{YOLO}~\cite{jocher2023yolo} and \textbf{SAM 2}~\cite{ravi2024sam} for real-time background removal. 
First, a lightweight YOLO detector, fine-tuned on a small set of real-world images, identifies the target object's bounding box. 
This box is then used as a spatial prompt for SAM 2 to generate a precise pixel-level mask in a zero-shot manner. 
Finally, the resulting binary mask is applied to the depth image to isolate the target geometry, serving as the clean observation for the control policy.
\subsubsection{Policy Deployment and Action Execution}
\paragraph{AdamU Deployment} 
For the real-robot deployment of AdamU, we implement an onboard asynchronous multi-threading mechanism across a heterogeneous computing platform, consisting of an NVIDIA Jetson Orin NX (high-level controller) and an Intel NUC (low-level controller).
\begin{itemize}
    \item \textit{Perception and Inference:} The ZED Mini camera is connected to the Orin NX, which serves as the unit for visual perception. The depth streams, acting as exteroceptive observations, are processed on the Orin NX. The policy, trained in simulation, is exported as an ONNX model for efficient deployment. Policy inference is executed on the Orin NX at approximately 1–2 Hz, predicting end-effector (EE) position increments and dexterous hand states (open/closed), which are then converted into target EE poses.
    \item \textit{Control and Feedback:} Data transmission between the Orin NX and Intel NUC is handled via ROS2. The Intel NUC receives the target EE poses, computes joint angles via the inverse kinematics (IK) solver, and commands the joint motors and dexterous hands at 100 Hz using a PID controller. Simultaneously, the NUC retrieves joint states, calculates actual EE poses via forward kinematics (FK), and streams this proprioceptive data back to the Orin NX at 100 Hz for closed-loop policy inference.
\end{itemize}

\paragraph{Airbot Deployment}
The Airbot Play also utilizes an asynchronous multi-threading framework, with all high-level processing centralized on an external RTX 4090 workstation.
\begin{itemize}
    \item \textit{Perception and Inference:} For this setup, the ZED Mini is connected to the RTX 4090 workstation. Proprioceptive states (EE poses) are retrieved at 200 Hz via the manufacturer's SDK. Similar to AdamU, the policy is deployed using the ONNX framework, performing inference at 2 Hz to generate EE position increments and target poses.
    \item \textit{Control and Feedback:} Control commands for the Airbot's EE poses and gripper states are sent directly through the manufacturer's SDK to the robot base, ensuring low-latency execution.
\end{itemize}

\FloatBarrier
\vspace{1cm}
\section{Cross-Simulator Experimental Setup}
We carefully tune baseline simulators to ensure numerical stability, maximize inextensibility, allow compliant bending, and avoid solver under-convergence. The specific physical properties and solver parameters for each simulator are detailed in Tables \ref{tab:params_flash}, \ref{tab:params_genesis}, \ref{tab:params_isaacsim}, and \ref{tab:params_newton}.

\begin{table}[tbh!]
    \centering
    \small
    \caption{Parameters of FLASH.}
    \label{tab:params_flash}
    \setlength{\tabcolsep}{20pt}
    \begin{tabular}{lc}
        \toprule
        \textbf{Cloth Physical Properties} & \\
        Bending stiffness & 0.2 \\
        Young's modulus & 3e4 \\
        Poisson's ratio & 0.4 \\
        \midrule
        \textbf{FLASH Solver Parameters} & \\
        Linear-solver iterations & 10 \\
        Local-global iterations & 5 \\
        \bottomrule
    \end{tabular}
\end{table}

\begin{table}[tbh!]
    \centering
    \small
    \caption{Parameters of Genesis.}
    \label{tab:params_genesis}
    \setlength{\tabcolsep}{20pt}
    \begin{tabular}{lc}
        \toprule
        \textbf{Cloth Physical Properties} & \\
        Bending compliance & 0.1 \\
        Stretch compliance & 1e-7 \\
        Bending relaxation & 0.1 \\
        Stretch relaxation & 0.3 \\
        Air resistance & 0 \\
        \midrule
        \textbf{PBD Solver Parameters} & \\
        Max. bending solver iterations & 10 \\
        Max. stretch solver iterations & 40 \\
        Max. density solver iterations & 10 \\
        Max. viscosity solver iterations & 10 \\
        \bottomrule
    \end{tabular}
\end{table}

\begin{table}[tbh!]
    \centering
    \small
    \caption{Parameters of Isaac Sim.}
    \label{tab:params_isaacsim}
    \setlength{\tabcolsep}{20pt}
    \begin{tabular}{lc}
        \toprule
        \textbf{Cloth Physical Properties} & \\
        Surface thickness & 1e-3 \\
        Surface bending stiffness & 0.2 \\
        Surface stretch stiffness & 1e3 \\
        Young's modulus & 3e4 \\
        Poisson's ratio & 0.4 \\
        \midrule
        \textbf{PGS Solver Parameters} & \\
        Min. position solver iterations & 100 \\
        Max. position solver iterations & 255 \\
        Min. velocity solver iterations & 100 \\
        Max. velocity solver iterations & 255 \\
        \bottomrule
    \end{tabular}
\end{table}

\begin{table}[tbh!]
    \centering
    \small
    \caption{Parameters of Newton.}
    \label{tab:params_newton}
    \setlength{\tabcolsep}{20pt}
    \begin{tabular}{lc}
        \toprule
        \textbf{Cloth Physical Properties} & \\
        Bending stiffness & 0.2 \\
        Bending damping & 0 \\
        Triangular elastic stiffness & 1e3 \\
        Triangular area stiffness & 1e3 \\
        Triangular damping stiffness & 1e-7 \\
        \midrule
        \textbf{VBD Solver Parameters} & \\
        Solver iterations & 100 \\
        \bottomrule
    \end{tabular}
\end{table}

\section{Bilateral Constraint for Cloth Control in FLASH}
We model the interaction between the virtual gripper and the cloth using position-level \textbf{bilateral constraints} \cite{2022SIGCOURSE}. Unlike penalty-based collision methods, bilateral constraints provide a stiff and stable attachment, effectively locking the selected cloth vertices to the gripper's trajectory.

Formally, let $\mathbf{x}_i \in \mathbb{R}^3$ denote the position of the $i$-th vertex of the cloth mesh, and $\mathbf{p}_{grip}(t)$ denote the target position of the virtual gripper at time $t$. For a set of grasped vertices $\mathcal{S}_{grip}$, we impose the following equality constraint for each vertex $i \in \mathcal{S}_{grip}$:
\begin{equation}
    \mathbf{C}(\mathbf{x}_i, t) = \mathbf{x}_i - \mathbf{p}_{grip}(t) = 0
\end{equation}
This constraint enforces that the relative distance between the cloth vertex and the gripper anchor remains zero, effectively creating a zero-length rigid link.

To simulate the grasping and releasing actions, we dynamically update the set $\mathcal{S}_{grip}$. When the gripper is activated, vertices within a proximity threshold are added to $\mathcal{S}_{grip}$ (constraint activation). Conversely, to release the fabric, the corresponding indices are discarded from $\mathcal{S}_{grip}$ (constraint deactivation), restoring the vertices' degrees of freedom governed solely by internal elastic forces and gravity.

\begin{figure*}[tbh!]
    \centering
    \begin{lstlisting}[
        language=C++, 
        basicstyle=\ttfamily\small, 
        frame=single
    ]
void BilateralizedSolver::setConstraintLinearSolverDense(linearsolver::dense::TypeDenseLinearSolver type)
{
    switch (type) {
        ...<linearsolver::DENSE_CHOLESKY_EIGEN>...
        ...<linearsolver::DENSE_CG>...
        case linearsolver::dense::DENSE_PCG_JACOBI:
        {
            _constraintlinearsolver = std::make_shared<linearsolver::iterative::DenseJacobiPCGSolver>(_maxIter, _tol, false);
            break;
        }
        case linearsolver::dense::DENSE_CR:
        {
            _constraintlinearsolver = std::make_shared<linearsolver::iterative::DenseCRSolver>(_maxIter, _tol, false);
            break;
        }
        ...<linearsolver::DENSE_PCR_JACOBI>...
    }
}
    \end{lstlisting}

    \vspace{-5pt}
    \caption{Code snippet demonstrating the initialization of our dense linear solver. The system supports dynamic switching between various algorithms (e.g., DENSE\_PCG\_JACOBI, DENSE\_CR) to strictly enforce bilateral constraints while maintaining numerical stability during complex gripper-cloth interactions.}
    \label{fig:bilateral_constraints}
\end{figure*}